%% file: main_aaai_medla.tex
\documentclass[letterpaper]{article} 
\usepackage{aaai2026}  
\usepackage{times}  
\usepackage{helvet}  
\usepackage{courier}  
\usepackage[hyphens]{url}  
\usepackage{graphicx} 
\usepackage{natbib}  
\usepackage{caption} 
\usepackage{algorithm} 
\usepackage{algorithmic}
\usepackage{amsmath} 
\usepackage{booktabs}
\usepackage{multirow}
\usepackage{array}
\usepackage{graphicx}
\usepackage{amssymb}
\usepackage{amsthm}
\newtheorem{property}{Property}

\theoremstyle{definition}
\newtheorem*{remark}{Remark}

\usepackage{newfloat}
\usepackage{listings}
\DeclareCaptionStyle{ruled}{labelfont=normalfont,labelsep=colon,strut=off} 
\floatstyle{ruled}
\newfloat{listing}{tb}{lst}{}
\floatname{listing}{Listing}
\title{MedLA: A Logic-Driven Multi-Agent Framework \\for Complex Medical Reasoning with Large Language Models}
\author{
    Siqi Ma\textsuperscript{\rm 1}\thanks{equal contribution},
    Jiajie Huang\textsuperscript{\rm 1$*$},
    Fan Zhang\textsuperscript{\rm 3}, 
    Yue Shen\textsuperscript{\rm 2$*$}, \\
    Jinlin Wu\textsuperscript{\rm 3},
    Guohui Fan\textsuperscript{\rm 4},
    Zhu Zhang\textsuperscript{\rm 4},
    Zelin Zang\textsuperscript{\rm 1,2,3}\thanks{corresponding author.}
}
\affiliations{
    \textsuperscript{\rm 1} Westlake University \\
    \textsuperscript{\rm 2} Ant Group\\
    \textsuperscript{\rm 3} CAIR, Hong Kong Institute of Science and Innovation (HKISI), CAS\\
    \textsuperscript{\rm 4} China-Japan Friendship Hospital \\
}

\usepackage{bibentry} 

\begin{document} 

\maketitle

\begin{abstract}
Answering complex medical questions requires not only domain expertise and patient-specific information, but also structured and multi-perspective reasoning. Existing multi-agent approaches often rely on fixed roles or shallow interaction prompts, limiting their ability to detect and resolve fine-grained logical inconsistencies. To address this, we propose \textsc{MedLA}, a logic-driven multi-agent framework built on large language models. Each agent organizes its reasoning process into an explicit logical tree based on syllogistic triads (major premise, minor premise, and conclusion), enabling transparent inference and premise-level alignment. Agents engage in a multi-round, graph-guided discussion to compare and iteratively refine their logic trees, achieving consensus through error correction and contradiction resolution. We demonstrate that \textsc{MedLA} consistently outperforms both static role-based systems and single-agent baselines on challenging benchmarks such as MedDDx and standard medical QA tasks. Furthermore, \textsc{MedLA} scales effectively across both open-source and commercial LLM backbones, achieving state-of-the-art performance and offering a generalizable paradigm for trustworthy medical reasoning.
\end{abstract}

\begin{links}
    \link{Code}{https://github.com/alexander2618/MedLA}
    \link{Extended version}{https://arxiv.org/abs/2509.23725}
\end{links}

%
\input{sec_intro.tex}
\input{sec_method.tex}

\input{sec_exp.tex}

\section{Conclusion} \label{Conclusion}
We present MedLA, a logic-driven multi-agent system that decomposes clinical questions into structured syllogistic trees and enables iterative, cross-agent discussions to resolve logical and knowledge conflicts. 
Notably, these improvements require no fine-tuning or external retrieval, highlighting the value of structured logic and collaborative reasoning.

\section{Acknowledgments}
This work was supported by the National Natural Science Foundation of China (No. 62306313), the Chinese Academy of Medical Sciences Innovation Fund for Medical Sciences (2024-I2M-TS-035), and the Zhejiang Province Selected Funding for Postdoctoral Research Projects (ZJ2025113). Additional support was provided by the InnoHK program, and Ant Group through the CAAI-Ant Research Fund. We also thank the Funding from the ROOTCLOUD TECHNOLOGY CO.,LTD.

We thank the China-Japan Friendship Hospital for providing medical expertise and resources, and Zhipu AI for their API support. We are grateful to Prof. Zhen Lei from the CAIR, Hong Kong Institute of Science and Innovation (HKISI) and Prof. Stan Z. Li from Westlake University for their valuable suggestions and comments.

\bibliography{aaai2026}

\appendix
\onecolumn

\input{app.tex}

\end{document}

%% file: sec_intro.tex
\section{Introduction}

\begin{figure*}[t]
    \centering
    \includegraphics[width=17.50cm]{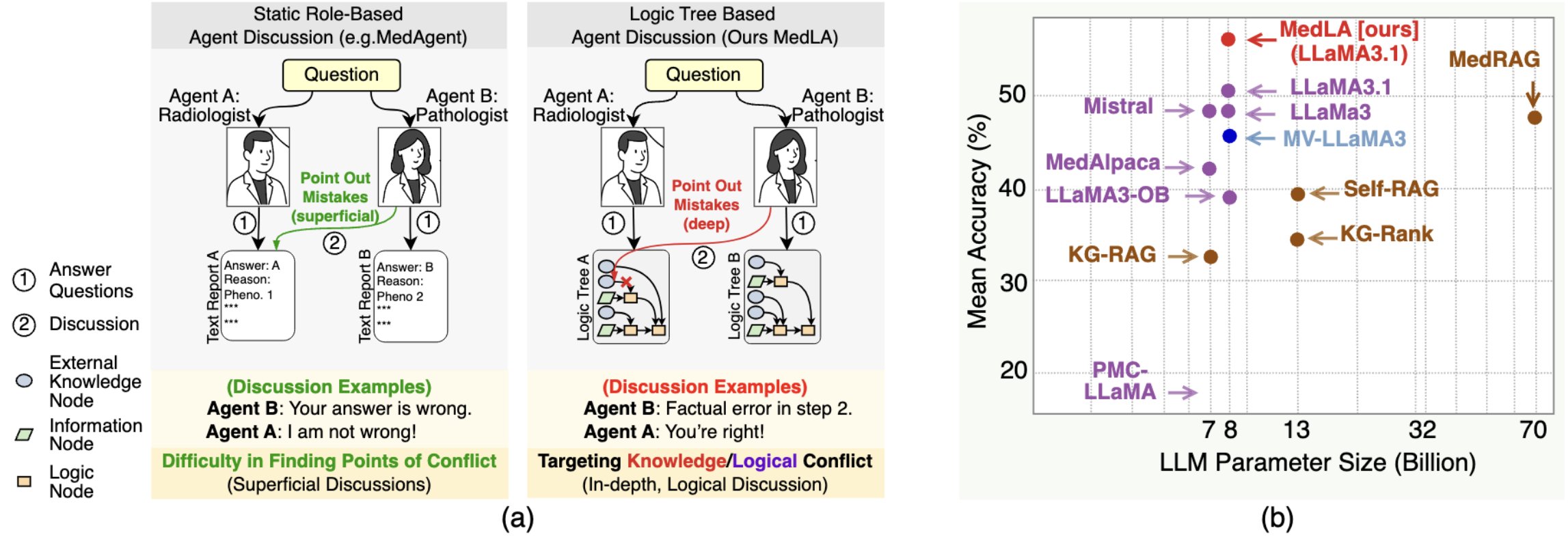}
    \caption{{(a) Comparison between traditional role-based agent discussions and our proposed logic-based framework. (b) Performance and parameter comparison of \textsc{MedLA} with existing systems.}
 (a-Left) Traditional systems (e.g., MedAgent) assign agents fixed roles and aggregate their conclusions, leading to superficial discussions and difficulty identifying the root of disagreement. (a-Right) Our approach models each agent's reasoning as a logic tree, enabling inter-agent analysis of logical and knowledge-based inconsistencies. (b) \textsc{MedLA} outperforms existing systems in the average accuracy of two benchmarks, demonstrating its effectiveness in handling complex medical reasoning tasks.}
    \label{fig_intro}
\end{figure*}

LLM has demonstrated significant advantages in the field of medical reasoning, and its deep learning architecture can efficiently extract knowledge from massive amounts of literature and clinical cases to provide intelligent assistance in diagnostic decision-making, and is expected to significantly improve the accessibility and popularization of medical knowledge\cite{chang2024survey,kasneci2023chatgpt}.
Answering complex medical questions with large language models (LLMs) remains difficult.  
A practical system must integrate domain knowledge~\citep{liu2024ddk}, patient information~\citep{yang2022large}, and explicit logical reasoning~\citep{goh2024large}.  
Recent general-purpose and domain-adapted models achieve strong open-benchmark scores~\citep{kim2024mdagents,tang2024medagents}.  
In clinical use, however, they may hallucinate drug dosages, misapply guidelines, or draw invalid causal links, reducing diagnostic reliability~\citep{liu2023llavamed}.

LLM-based medical reasoning follows two main paradigms.  
\emph{(a) Knowledge fine-tuning~\citep{singhal2023large,wang2025knowledge}} retrains models on large medical corpora, improving accuracy at the cost of data, compute, and deployment agility.
\emph{(b) Reasoning stimulation~\citep{lievin2024can}} has been tried using multi-agent role-playing to accomplish tasks through discussion and cooperation, which is considered a flexible and low-cost solution~\citep{moor2023medflamingo,kim_mdagents_2024}. However, we found that most current multi-agent systems only engage in positional discussions based on their rulings and cannot argue about logical details in depth. Current frameworks are not able to effectively localize logic/rule conflicts, thus be difficult to improve the performance. 

Inspired by the `major premise-minor premise-conclusion' paradigm of the classical syllogism, we use the syllogism as a minimal reasoning unit ~\citep{khemlani2012theories,jiang2023legal}. Each triad consists of a generalized medical law (major premise), a patient-specific fact (minor premise), and the corresponding conclusion. By concatenating or parallelizing multiple triads, we construct an inference tree whose leaf nodes store empirical observations or domain rules, internal nodes hold intermediate inferences, and the root node gives the final clinical decision ~\citep{howson2005logic}. Such inference trees offer two major advantages:
(i) Traceability - each conclusion can be traced back to its supporting premises;
(ii) Comparability - each intelligence can align the reasoning tree~(logical tree) to pinpoint conflicts or omissions at the premise level.
Embedding this explicit structure into a multi-agent framework provides an easily auditable logical template for systematic cross-intelligence error correction and consistency verification.

To this end, we propose a medical logical tree-based multi-agent framework~(termed \textsc{MedLA}) that enables multi-agent collaboration and discussion through a logic tree structure~(in Fig.~\ref{fig_intro}(a)). \emph{Although LLM cannot explicitly handle tree logic, we can organize the answers to medical questions in the form of a logic tree by guiding them with prompt words. }We design premise agents to extract the major premise and minor premise from the questions. Sub-questions are generated by the premise agents and sent to the logical agents. Then we use medical agents to reason about the logical nodes and generate the logical tree with the final answer. Based on the logical tree, a multi-round discussion mechanism is designed to allow agents to discuss the logical tree and correct each other's errors. The final answer is generated based on the logical tree and the discussion results.

The experimental results show that our MedLA outperforms existing MedDDx benchmarks~\citep{su2025kgarevion}, medical QA benchmarks~\citep{xiong2024medrag}, and medical reasoning benchmarks~\citep{zuo2025medxpertqa} by a large margin on both open-source and commercial LLM~(Fig.~\ref{fig_intro}(b)).
\textbf{Contributions.}  
(a) We introduce \textsc{MedLA}, the first multi-agent framework for medical reasoning that represents each agent's thought process as an explicit logical tree. This design enables fine-grained traceability of inferences and systematic detection of premise-level conflicts.
(b) We develop a multi-round, graph-guided discussion mechanism in which agents iteratively compare and revise their logical trees, leading to robust cross-agent error correction and convergence to high-confidence, self-consistent reasoning structures.
(c) We perform comprehensive evaluations on both differential diagnosis (\textsc{MedDDx}) and standard medical QA benchmarks, demonstrating that \textsc{MedLA} outperforms static role-based multi-agent systems and single LLM baselines.

%% file: sec_method.tex
\section{Methods} \label{Methods}

\begin{figure*}
   \centering
   \includegraphics[width=17.00cm]{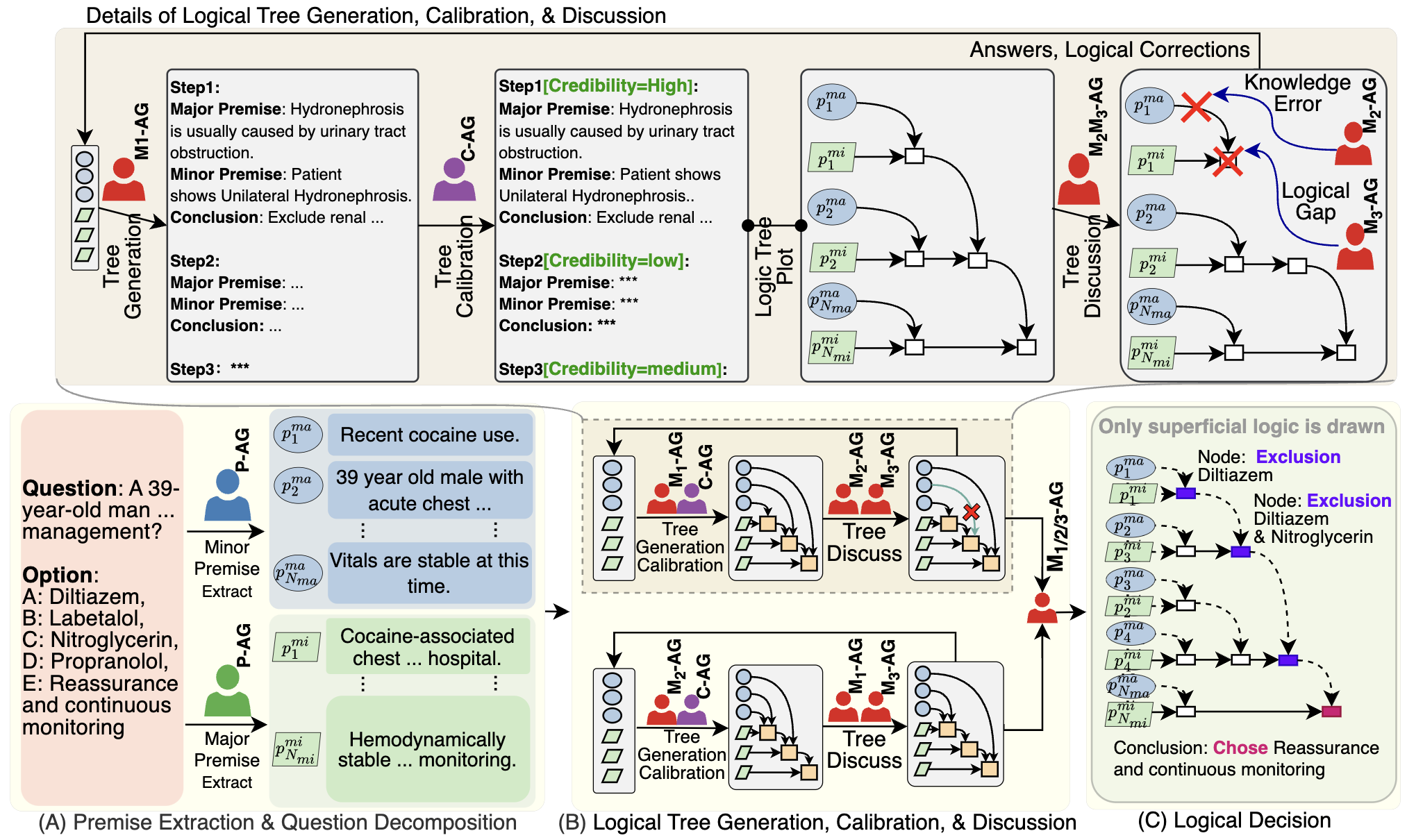}
   \caption{{Overview of the proposed MedLA for complex medical reasoning.} The system decomposes a medical query into logical sub-tasks, dynamically invokes specialized agents, and engages in collaborative reasoning to generate answers.
   }
    \label{fig:overview} 
\end{figure*}

\subsection{Problem Definition, Syllogism, and Logical Tree}
In this work, each clinical QA task dataset $\mathcal D$ is defined as a set of tuples $\{(Q_i, \mathcal O_i, C_i)\}_{i=1}^N$, where $Q_i=\text{`What is the correct diagnosis/management for the patient?'}$ is the {question text}, $\mathcal O_i =\{O_{i_1},O_{i_2},\dots,O_{i_{N_K}}\}$ is the set of {candidate answers}, $N_K$ is the number of candidate answers, and $C_i \in \mathcal O_i$ is the ground-truth {correct answer}. The task for our system is to predict the correct answer from $\mathcal O_i$ given the input $(Q_i, \mathcal O_i)$. We primarily consider the {Multiple-choice Question (MCQ)} format~\citep{yun2023medmcqa}.

\paragraph{Syllogism and Syllogism-based logical tree.}  A minimal reasoning unit in medical diagnosis can be abstracted as the classical syllogism~\citep{smiley1973syllogism}, 
  $ v: (p^\text{maj} \text{ - major premise}) \;\wedge\; (p^\text{min} \text{ - minor premise}) \\
   \quad \rightarrow (C \text{ - conclusion}), $
where $v$ is the syllogism node, which is a tuple of the {major premise} $p^\text{maj}$, the {minor premise} $p^\text{min}$, and the {conclusion} $C$.
By chaining or paralleling multiple syllogisms, we obtain a logical tree~\citep{khemlani2012theories,revlis2015syllogistic},
\begin{equation}
\begin{split}
    \mathcal T & =(V, E), E  \subseteq V\times V, \\
    V & =\{v_1, v_2, \cdots, v_i, \cdots,v_{N_K}\}, \\
\end{split}
\label{eq_syllogism}
\end{equation}
where $\mathcal T$ denotes the syllogism-based logical tree, $V$ is the set of syllogism nodes,$v_i$ is defined in Eq.~(\ref{eq_syllogism}), and $E$ is the set of directed edges. Each $(v_{i_1}, v_{i_2}) \in E$ means that $v_{i_1}$ is a necessary antecedent of $v_{i_2}$. By constructing a logical tree, we can represent the reasoning process in a structured manner, allowing for better understanding and analysis of the reasoning steps involved in concluding.We further establish the theoretical properties of this reasoning framework, such as its convergence and stability, with proofs provided in the Appendix.

\subsection{Agent Designs in MedLA}
\label{sec:agent_design}
Based on the above definitions of logical tree, we propose a multi-agent framework for complex medical reasoning, called \textsc{MedLA}, which is designed to handle complex medical questions by dynamically invoking specialized agents to collaboratively reason and generate comprehensive answers~(in Fig.~\ref{fig:overview}). The proposed framework consists of four main components, including a {Premise Agent} (P-Agent) for major/minor premise extraction, a {Decompose Agent} (D-Agent) for question splitting, multiple {Medical Agents} (M-Agents) for recursive tree generation, and a {Credibility Agent} (C-Agent) for node calibration. The system is designed to be modular and flexible, allowing for dynamic task decomposition and multi-agent collaboration.

\paragraph{Premise Agent (P-Agent) for major/minor premise extraction.}
The first step maps free text $Q$ and external knowledge into {major/minor premises} that seed later syllogisms. From the question $Q$ we extract entity-relation facts to obtain a patient fact set $\mathcal P^{\text{min}}$, and retrieve relevant rules $\mathcal P^{\text{maj}}$ from medical knowledge bases,
\begin{equation}
  \begin{aligned}
    \mathcal P^{\text{maj}}=\{p^{\text{maj}}_1,\dots, p^{\text{maj}}_r,\dots, p^{\text{maj}}_{|N_\text{maj}|}\}=\mathcal O_{\mathrm{P\text{-}Agent}}^{\langle\texttt{pro}_{P_\text{maj}}\rangle}(Q), \\
    \mathcal P^{\text{min}}=\{p^{\text{min}}_1, \dots, p^{\text{min}}_r,\dots,p^{\text{min}}_{|N_\text{min}|}\}=\mathcal O_{\mathrm{P\text{-}Agent}}^{\langle\texttt{pro}_{P_\text{min}}\rangle}(Q),
  \end{aligned}
\end{equation}
where $\mathcal P^{Q,k}$ is the {knowledge-major-premise set}, $\mathcal P^{Q,\text{min}}$ is the {patient-minor-premise set}; each $p^{\text{maj},(n)}$ denotes the $n$-th retrieved medical rule, and each $p^{\text{min},(m)}$ denotes the $m$-th patient fact extracted from $Q$, $\mathcal O_{\mathrm{P\text{-}Agent}}(\cdot)$ is the P-Agent function, and $\langle\texttt{pro}_{P_\text{maj}}\rangle$ and $\langle\texttt{pro}_{P_\text{min}}\rangle$ are the prompt templates. $|N_\text{maj}^{Q}|$ and $|N^{Q}_\text{min}|$ are the number of major and minor premises, respectively. The fixed premise extract prompt is listed in the Appendix.

\paragraph{Decompose Agent (D-Agent) for question splitting.}
Complex diagnostic problems often span multiple causal chains; decomposing these problems into atomic subproblems allows for more comprehensive thinking and discussion. D-Agent recursively splits $Q$ into item questions $\{q_1,q_2,\dots\}$,
\begin{equation}
  \mathcal S = \{s_1,s_2,\dots\}
  =\mathcal O_{\mathrm{D\text{-}Agent}}^{\langle\texttt{pro}_{D}\rangle}
    \!\bigl(Q\bigr),
\end{equation}

where $\mathcal O_{\mathrm{D\text{-}Agent}}(\cdot)$ is the D-Agent function and $\langle\texttt{pro}_{D}\rangle$ is the prompt template~(in the Appendix). 
The D-Agent prompt is designed to elicit a tree-like structure of questions, where each question $q_s$ is a subproblem that can be answered independently. In this work, we adopt an {elimination-based reasoning strategy} to construct the question tree: if candidate answers are provided, the agent considers the plausibility of each option in turn using a process of elimination; for open-ended questions, the agent is first prompted to generate multiple possible answers and then considers each as an independent hypothesis.

\paragraph{Medical Agents (M-Agents) for logical tree generation.}
To obtain {diverse and complementary} reasoning perspectives, we run multiple M-Agents $ \mathcal M = \{M^{(1)},\dots, M^{(j)}, \dots,M^{(N_M)},\}$ in parallel, where $N_M$ is the number of M-Agents. Agent $M^{(j)}$ independently generates the next batch of derivable conclusions,
\begin{equation}
  \mathcal T_{M^{(j)}} = \{V, E\}
=\mathcal O_{\mathrm{M^{(j)}\text{-}Agent}}^{\langle\texttt{pro}_{M}\rangle}
  \bigl(\mathcal P^{\text{maj}},\,\mathcal P^{\text{min}},\mathcal S, {\mathcal T}_\text{other}\bigr), 
\end{equation}
where $\mathcal O_{\mathrm{M^{(j)}\text{-}Agent}}(\cdot)$ is $j$-th M-Agent, and 
$\langle\texttt{pro}_{M}\rangle$ is the prompt template~(in the Appendix). 
The prompt will guide the model to further split the subproblems and thus form a more complex logical tree $\mathcal T_{M^{(j)}}$.
$\mathcal T_{M^{(j)}}$ contains a set of syllogism nodes $V=\{v_1, v_2, \dots, v_i, \dots, v_{|N_V|}\}$ and a set of directed edges $E$. 
${\mathcal T}_\text{other}$ is an optional input, which is used to provide the tree structure of other M-Agents in the previous round for agent-agent discussion. 
It is worth noting that in this paper, since LLMs cannot directly produce structured tree outputs, the $v_i$ we obtain by guiding LLM is a text block containing a syllogism(check Appendix).

\paragraph{Credibility Agent (C-Agent) for node calibration.}
To ensure the credibility of the generated logical tree, we introduce a C-Agent to evaluate the confidence of each syllogism node. The C-Agent is designed to assess the logical consistency and relevance of each node in the tree, providing a credibility score for each syllogism.
\begin{equation}
  \mathcal C_{M^{(j)}} =\{c_1, c_2, \dots, c_i, \dots, c_{|N_V|}\} 
     =\mathcal O_{\mathrm{C\text{-}Agent}}^{\langle\texttt{pro}_{C}\rangle}
       \bigl(T_{M^{(j)}}\bigr), 
\end{equation}
where $\mathcal O_{\mathrm{C\text{-}Agent}}(\cdot)$ is the function of C-Agent, and $\langle\texttt{pro}_{C}\rangle$ is the prompt template~(in the Appendix). In this paper, we define $c_i\in\{\text{High credibility, Medium credibility, Low credibility}\}$, where $i$ is the index of the syllogism node $v_i$ in the logical tree $\mathcal T_{M^{(j)}}$. 
\subsection{Logical Reasoning Workflow of MedLA} \label{sec:logical_tree}

Based on the above agent designs, we propose a three-stage pipeline for logical reasoning in complex medical QA tasks. 

\textbf{Phase A: Premise Extraction \& Question Decomposition.}  
The system begins with the P-Agent: given the raw question $Q$, it uses fixed prompt templates to extract a set of knowledge-major premises $\mathcal P^{\mathrm{major}}$ and patient-minor premises $\mathcal P^{\mathrm{minor}}$. Next, the D-Agent recursively splits $Q$ into atomic sub-questions $\{q_1,q_2,\dots\}$, creating placeholder nodes for each. If a sub-question coincides with a candidate answer, it is marked as a terminal goal; otherwise, it remains pending for further inference.

\textbf{Phase B: Logical Tree Generation, Calibration, \& Discussion.}  
A cohort of M-Agents runs in parallel. Each M-Agent independently takes $\mathcal P^{\mathrm{major}}$, $\mathcal P^{\mathrm{minor}}$, and the set of placeholders, then emits a batch of syllogistic nodes and directed edges in one LLM call, forming a provisional local logical tree. Subsequently, the C-Agent reevaluates each node's confidence. Nodes labeled as low confidence are flagged and retained as discussion material for the next phase; all medium-high confidence nodes are locked into the local tree to preserve core reasoning structure. Then the system enters a discussion phase, where each M-Agent exchanges its local tree with others, allowing them to compare and contrast reasoning paths. These discussions repeat until all agents have shared their trees. The system then identifies any discrepancies between the trees, focusing on the flagged low-confidence nodes. Each agent is prompted to review and revise its tree based on the feedback from its peers, using a revision prompt ($\langle\texttt{pro}_{\mathrm{Rev}}\rangle$) to validate, add or remove premises, and re-score affected nodes.

\input{tab_medddx.tex}

\textbf{Phase C: Logical Decision.}  
Once the discussion phase is complete, the system synthesizes the final logical tree by merging all local trees. The final tree is then used to generate the final answer. The answer is generated by traversing the logical tree and aggregating the conclusions from each syllogism node. The system can also provide a detailed explanation of the reasoning process, including the major and minor premises, the syllogisms used, and the final conclusion.

%% file: tab_medddx.tex
\begin{table*}[t]
    \centering
    
        \footnotesize
        \begin{tabular}{c|l|c||ccc|c}
            \toprule
                                                                                                                  & \multirow{3}{*}{\textbf{Method}} & \multirow{3}{*}{\textbf{Reference}} & \multicolumn{4}{c}{\textbf{MedDDx Benchmarks \citep{su2025kgarevion}}}                                                                                             \\
                                                                                                                  &                                  &                                     & \textbf{Basic}                                                         & \textbf{Intermediate}      & \textbf{Expert}             & \multirow{2}{*}{\textbf{AVE}}  \\
                                                                                                                  &                                  &                                     & Acc.($\pm$std)                                                         & Acc.($\pm$std)             & Acc.($\pm$std)              &                                \\ \midrule
            \multirow{3}{*}{{\begin{tabular}[c]{@{}c@{}}Graph \\Based\\ Methods\end{tabular}}}                    & QAGNN                            & NAACL2021                           & \underline{29.5($\pm$0.3)}                                             & \underline{26.5($\pm$0.2)} & \underline{25.3($\pm$0.3)}  & \underline{27.1}               \\
                                                                                                                  & JointLK                          & NAACL2022                           & 24.7($\pm$0.4)                                                         & 25.3($\pm$0.4)             & 24.4($\pm$0.4)              & 24.8                           \\
                                                                                                                  & Dragon                           & NeurIPS2022                         & 28.6($\pm$0.3)                                                         & 24.7($\pm$0.2)             & 24.0($\pm$0.4)              & 25.8                           \\ \midrule
            \multirow{4}{*}{{\begin{tabular}[c]{@{}c@{}}Multi \\ Agents \\ Methods\end{tabular}}}                 & MV-LLaMA3.1(8B)                  & -                                   & 39.6($\pm$1.0)                                                         & 32.8($\pm$0.6)             & 30.2($\pm$0.8)              & 34.2                           \\
                                                                                                                  & DyLAN                            & COLM2024                            & 39.3($\pm$1.5)                                                         & 33.5($\pm$0.9)             & 31.1($\pm$0.7)              & 34.6                           \\
                                                                                                                  & MedAgents                        & ACL2024                             & 41.0($\pm$0.7)                                                         & 35.7($\pm$1.1)             & 32.9($\pm$1.5)              & 36.5                           \\
                                                                                                                  & MDAgents                         & NeurIPS2024                         & \underline{42.1($\pm$1.3)}                                             & \underline{37.5($\pm$0.9)} & \underline{33.4($\pm$0.6)}  & \underline{37.7}               \\ \midrule
            \multirow{5}{*}{{\begin{tabular}[c]{@{}c@{}}General \\\&\\ Medical\\ LLMs\end{tabular}}}                                                                                                                    & Mistral(7B)                      & Mistral2023                         & 41.2($\pm$0.3)                                                         & 35.6($\pm$0.3)             & \underline{37.5($\pm$0.7)}  & \underline{38.1}               \\
                                                                                                                  & MedAlpaca(7B)                    & BHT2023                             & 39.9($\pm$1.2)                                                         & 32.5($\pm$0.4)             & 31.1($\pm$0.9)              & 34.5                           \\
                                                                                                                  & PMC-LLaMA(7B)                    & SJTU2024                            & 8.7($\pm$1.5)                                                          & 8.6($\pm$0.2)              & 7.9($\pm$0.6)               & 8.4                            \\
                                                                                                                  & LLaMA3(8B)                       & Meta2024                            & 42.8($\pm$0.5)                                                         & 31.9($\pm$0.2)             & 30.6($\pm$0.9)              & 35.1                           \\
                                                                                                                  & \textbf{LLaMA3.1(8B)[baseline]}  & Meta2024                            & \underline{43.4($\pm$1.8)}                                             & \underline{36.8($\pm$0.2)} & 30.6($\pm$2.1)              & 36.9                           \\ \midrule
            \multirow{5}{*}{{\begin{tabular}[c]{@{}c@{}}General \\\&\\ Medical\\ LLMs \\with COT\end{tabular}}}                                                                                                       & CoT-Mistral(7B)                  & Mistral2023                         & 40.4($\pm$1.0)                                                         & 36.8($\pm$2.3)             & \underline{37.9($\pm$2.7)}  & 38.4                           \\
                                                                                                                  & CoT-MedAlpaca(7B)                & BHT2023                             & 39.5($\pm$0.7)                                                         & 32.1($\pm$1.1)             & 31.2($\pm$1.0)              & 34.3                           \\
                                                                                                                  & CoT-PMC-LLaMA(7B)                & SJTU2024                            & 8.8($\pm$0.2)                                                          & 7.7($\pm$0.4)              & 6.3($\pm$0.5)               & 7.6                            \\
                                                                                                                  & CoT-LLaMA3(8B)                   & Meta2024                            & \underline{43.4($\pm$0.9)}                                             & 36.8($\pm$0.4)             & 31.3($\pm$0.3)              & 37.2                           \\
                                                                                                                  & CoT-LLaMA3.1(8B)                 & Meta2024                            & {43.9($\pm$1.7)}                                                       & \underline{39.3($\pm$0.5)} & 32.2($\pm$1.4)              & \underline{ 38.5 }             \\ \midrule
            \multirow{4}{*}{{\begin{tabular}[c]{@{}c@{}}RAG \\\&\\ Based\\ Methods\end{tabular}}}                 & Self-RAG(7B)                     & ICLR2024                            & 23.8($\pm$0.7)                                                         & 19.9($\pm$3.7)             & 22.4($\pm$4.5)              & 22.0                           \\
                                                                                                                  & Self-RAG(13B)                    & ICLR2024                            & 24.9($\pm$1.0)                                                         & 29.0($\pm$1.8)             & 26.6($\pm$3.1)              & 26.8                           \\
                                                                                                                  & KG-Rank(13B)                     & ACL-w2024                           & 25.3($\pm$2.1)                                                         & 25.6($\pm$1.3)             & 23.4($\pm$1.0)              & 24.8                           \\
                                                                                                                  & MedRAG(70B)                      & Oxon2024                            & \underline{36.5($\pm$0.8)}                                             & \underline{34.8($\pm$1.1)} & \underline{ 32.7($\pm$0.3)} & \underline{ 34.7 }             \\ \midrule
            \multirow{1}{*}{{\begin{tabular}[c]{@{}c@{}}Logic Based\end{tabular}}}                                & MedLA+LLaMA3.1(8B)               & Ours                                & \textbf{48.2($\pm$1.2)}                                                & \textbf{43.0($\pm$2.1)}    & \textbf{41.7($\pm$0.8)}     & \textbf{44.3 ($\uparrow$ 7.4)} \\
            \bottomrule
        \end{tabular}
      \caption{{The performance of our MedLA on MedDDx Benchmarks.} The table includes the accuracy along with the standard deviation($\pm$std) for each metric. The results demonstrate the effectiveness of MedLA in addressing complex medical reasoning tasks across different datasets. OB means OpenBioLLM, MV means Majority Voting, and AVE means averaged accuracy. The best results are highlighted in {bold}, while the best results of each method's block are marked with an \underline{underline}. Reference indicates the paper where the method was first introduced. The results of the baselines are taken from \citep{su2025kgarevion}.}\label{tab_main_results_medddx1}
\end{table*}

%% file: sec_exp.tex
\section{Experiments} \label{Experiments}

\textbf{Datasets \& Benchmarks.}  
We evaluate on three complementary benchmarks.
\noindent\textbf{(i) {MedDDx benchmarks}.}  
To stress differential-diagnosis reasoning, we use {MedDDx benchmark\citep{su2025kgarevion} }:  Tests differential diagnosis across \textbf{Basic}, \textbf{Intermediate}, and \textbf{Expert} tiers, with difficulty defined by the semantic similarity of distractors from STaRK-Prime \citep{wu2024stark}. 
\noindent\textbf{(ii) {Multi-choice medical QA benchmarks}} \citep{xiong2024medrag}.  
A suite combining \textbf{MMLU-Med}, \textbf{MedQA-US}, and \textbf{BioASQ-Y/N} to test general medical knowledge. 
It covers factual recall, guideline interpretation, and clinical decision-making.
\noindent\textbf{(iii) {Expert-Level Medical Reasoning and  Understanding benchmark}}\citep{zuo2025medxpertqa}.To assess expert-level reasoning, we use \textbf{MedXpertQA}, a challenging and comprehensive benchmark for advanced medical knowledge (details in Appendix).  
The benchmark serves as our {general-purpose} test bed. The details of the benchmarks are shown in Appendix.

\input{tab_medqa.tex}
\textbf{Baselines.}  
We benchmark {MedLA} against four representative paradigms.  
(i) \textbf{Graph-based reasoning} methods ground answers on biomedical knowledge graphs—{QAGNN} \citep{yasunaga-etal-2021-qa}, {JointLK} \citep{sun-etal-2022-jointlk}, and {DRAGON} \citep{yasunaga2022dragon}.  
(ii) \textbf{Multi-agent voting} systems aggregate LLM outputs without explicit logic trees—{Majority Voting}, {DyLAN} \citep{liu2024dynamicllmpoweredagentnetwork}, {MedAgents} \citep{tang2024medagents}, and {MDAgents} \citep{liu2024dynamic}.  
(iii) \textbf{Stand-alone LLMs} rely purely on parametric knowledge, including {LLaMA-2}-7B/13B, {Mistral-7B}, {MedAlpaca-7B}, {PMC-LLaMA-7B}, {LLaMA 3}-8B, {LLaMA 3-OB}-8B, and the stronger {LLaMA 3.1-8B}; we also report their chain-of-thought (CoT) variants.  
(iv) \textbf{Retrieval-augmented generation (RAG)} couples an LLM with external retrievers—{Self-RAG} (7B/13B) \citep{asai2023selfrag}, {KG-Rank} \citep{yang2024kgrank}, {KG-RAG} \citep{soman2023kgrag}, and {MedRAG} (70B) \citep{xiong2024medrag}.
Together, these baselines span parametric, retrieval-augmented, graph-grounded, and naïve multi-agent strategies, furnishing a comprehensive backdrop against which to gauge MedLA's contributions. We did not include methods that require additional data and require fine-tuning of the larger model~(e.g., KGAREVION~\citep{su_kgarevion_2025}).

\textbf{Evaluation Metric \& Testing Protocol \& Implementation Details.}  
Following \citep{su2025kgarevion}, the performance is measured by {accuracy} (Acc), averaged over three independent runs (\,$\pm$\,std).  
For every run, we shuffle the benchmark order and reset the LLM's sampling state.  
All experiments used the officially provided base model weights and configurations, and vLLM (v0.7.2).  8-card A100-80GB GPU servers are used for testing. The raw data of the dataset involved in the experiments is adopted from MIRAGE\footnote{\url{https://github.com/Teddy-XiongGZ/MIRAGE}}. More details can be found in the Appendix.

\textbf{[Overall Performance Analysis] MedLA significantly outperforms baselines under open-source LLM settings.}
To evaluate the effectiveness of MedLA, we conducted comprehensive experiments on a series of medical reasoning benchmarks, all based on open-source large language models (e.g., LLaMA). To ensure the objectivity of the comparison, we directly refer to the baseline results reported in \citep{su2025kgarevion}. We also provide a detailed summary in the Appendix.

\textbf{Analysis:}
(a) MedLA outperforms all baselines on both standard and challenging benchmarks, achieving state-of-the-art (SOTA) results across QA datasets and excelling in expert-level diagnosis on MedDDx. These results demonstrate that MedLA not only enhances factual reasoning but also improves differential diagnostic performance in real-world medical settings.
(b) {MedLA Beyond CoT and Baselines:} MedLA surpasses both base and CoT-enhanced models under the same open-source foundation, showing that our logic extraction strategy effectively distills more reliable knowledge and supports dynamic reasoning.
(c) {MedLA Stronger than Multi-Agent Systems:} MedLA outperforms multi-agent baselines, indicating that its logic tree enables deeper interaction and better coordination among agents even without explicit role-based decomposition.
(d) {MedLA Outperforms RAG without External Knowledge:} MedLA exceeds RAG-based models despite not using external retrieval, proving its strong internal reasoning ability through structured logic alone.
(e) {MedLA Remains Effective Under New Challenges:}On a newly introduced test set,MedXpertQA, MedLA's performance remains outstanding, proving that its logic-enhanced reasoning ability is generalizable, rather than merely an optimization for known tasks, and is capable of effectively tackling new medical challenges. 

\textbf{[Overall Performance Analysis] MedLA demonstrates strong performance advantages on commercial LLMs, validating its generality and robustness.}
Our previous experiments were primarily conducted on open-source large language models, where MedLA had already shown promising results across various medical reasoning tasks. To further assess the adaptability and competitiveness of MedLA under more powerful backbones, we conducted additional evaluations using state-of-the-art commercial LLMs such as DeepSeek. All baseline methods were also implemented on the same commercial model to ensure fair comparison. For objectivity, we referenced the baseline results reported in \citep{su2025kgarevion}. The outcomes are presented in Table~\ref{tab:deepseek_medical_reasoning}.

\input{tab_expert.tex}

\input{tab_smaltab.tex}
\input{fig_small.tex}

\textbf{[Detailed Difficulty vs. Performance Analysis] MedLA gives greater lift to more difficult tasks.}
We examine how the proposed logic-tree framework scales with task difficulty by comparing MedLA to its LLaMA-3.1-8B backbone on the three graded subsets of MedDDx (basic, intermediate, expert). Three random seeds are evaluated per tier; mean accuracy and standard deviation are reported in Fig.~\ref{fig_MedDDx}-left.  All experimental factors—retriever, decoding temperature, and candidate pool—are controlled, so any performance delta reflects the contribution of MedLA's multi-agent reasoning.

\textbf{Analysis:}
The relative improvement grows monotonically with difficulty: +4.6 pp~(percentage point) on basic, +6.4 pp on intermediate, and +11.1 pp on the expert tier.  Confidence intervals also narrow, indicating increased prediction stability.  These results suggest that explicit logic-tree exchange and cross-agent revision become progressively more beneficial as diagnostic options converge semantically, reinforcing the value of structure-level reasoning for the most challenging clinical cases.

\textbf{[Detailed Base Model vs. Performance Analysis] MedLA Continuous Enhancement on a Stronger Base Model.}
To assess MedLA's benefits beyond a single backbone, we compare its gains over both the 8-bit and 70-bit variants of LLaMA-3.1 on the MedDDx-Expert tier~(in Fig.~\ref{fig_MedDDx}-right). 

\textbf{Analysis:} MedLA yields consistent, substantial improvements on both backbones: from 30.6\% to 41.7\% (+11.1 pp) for 8B, and from 41.8\% to 51.9\% (+10.1 pp) for 70 B. These results demonstrate that our structured, multi-agent reasoning adds value even as the underlying LLM scales up, underscoring the generality and robustness of the logic-tree approach across model sizes.


\textbf{[Ablation Study] Each MedLA module contributes additively to final accuracy.}
To evaluate the independent contributions of the three modules of logic tree, confidence calibration, and multi-round correction, we eliminated each component one by one under the same LLaMA-3.1-8B backbone, the same cueing and decoding hyperparameters (see Table~\ref{tab_ablation}), and averaged them with 3 random seeds.

\textbf{Analysis:}
Dropping the {Revision loop} lowers accuracy by 2.2 pp on {MedQA-US} and roughly 1.8 pp on both MedDDx tiers, confirming that structured peer feedback yields tangible gains. Suppressing the {Credibility Agent} causes an additional 1.1-1.4 pp decline, showing that calibrated confidence scores steer agents towards more reliable updates.
Removing the entire {Logic-tree scaffold}—thereby falling back to plain chain-of-thought outputs—produces the steepest drop (-4.5 pp on {MedQA-US}, -4.3 pp on MedDDx-Expert).



\textbf{[Time Consumption Analysis] The complexity and time consumption of MedLA is manageable.}
To quantify the latency of the different methods in a real inference process, we recorded the wall-clock time on the \textsc{BioASQ-Y/N} dataset. All models are deployed on the same A100-80GB, and the decoding temperature is kept consistent with the number of concurrent threads to avoid interference from hardware differences. Table \ref{tab_time} splits the total elapsed time into four components: additional fine-tuning (FT), external retrieval (RT), logic graph construction and revision (GBT), and pure language model inference (IFT). For KGAREVION, which is only parameter fine-tuning, we add the officially reported 10 k-second training time to the total cost. 

\textbf{Analysis:}
Most baselines perform forward inference only once, with latency linearly proportional to model size; KGAREVION performs rapid inference but requires several additional hours of fine-tuning and has the highest overall cost. medLA performs stepwise inference through 17 subagents, which is about 2 times higher than the simple majority-voting scheme but much lower than KGAREVION, which requires offline training and is within the acceptable range. within the acceptable interval. More importantly, MedLA does not introduce additional retrieval or offline fine-tuning sessions.

%% file: tab_medqa.tex
\begin{table*}[t]
  \centering
 
  \footnotesize
  \begin{tabular}{c|l|c||ccc|c} \toprule
                                                                                                         & \multirow{3}{*}{\textbf{Method}} & \multirow{3}{*}{\textbf{Reference}} & \multicolumn{4}{c}{\textbf{Multi-choice medical QA benchmarks \citep{xiong2024medrag}}}                                                                                             \\
                                                                                                         &                                  &                                     & MMLU-Med                                                                                & MedQA-US                    & BioASQ-Y/N                  & \multirow{2}{*}{\textbf{AVE}} \\
                                                                                                         &                                  &                                     & Acc.($\pm$std)                                                                          & Acc.($\pm$std)              & Acc.($\pm$std)              &                               \\ \midrule
    \multirow{3}{*}{{\begin{tabular}[c]{@{}c@{}}Graph \\ Based \\ Methods\end{tabular}}}                 & QAGNN                            & NAACL2021                           & 31.7($\pm$0.6)                                                                          & 47.0($\pm$0.3)              & \underline{70.7($\pm$0.6)}  & 49.8                          \\
                                                                                                         & JointLK                          & NAACL2022                           & 28.8($\pm$0.6)                                                                          & 42.5($\pm$0.2)              & 70.6($\pm$0.5)              & 47.3                          \\
                                                                                                         & Dragon                           & NeurIPS2022                         & \underline{31.9($\pm$0.3)}                                                              & \underline{47.5($\pm$0.2)}  & 70.6($\pm$0.3)              & \underline{50.0}              \\ \midrule
    \multirow{4}{*}{{\begin{tabular}[c]{@{}c@{}}Multi \\ Agents \\ Methods\end{tabular}}}                & MV-LLaMA3.1(8B)                  & -                                   & 60.2($\pm$0.5)                                                                          & 46.8($\pm$0.4)              & \underline{65.2($\pm$0.4)}  & 57.4                          \\
                                                                                                         & DyLAN                            & COLM2024                            & 62.5($\pm$0.3)                                                                          & 51.6($\pm$0.6)              & 63.8($\pm$0.5)              & 59.3                          \\
                                                                                                         & MedAgents                        & ACL2024                             & 64.3($\pm$0.4)                                                                          & 53.2($\pm$0.3)              & 64.1($\pm$0.6)              & 60.5                          \\
                                                                                                         & MDAgents                         & NeurIPS2024                         & \underline{65.0($\pm$0.2)}                                                              & \underline{53.4($\pm$0.2)}  & 64.0($\pm$0.3)              & \underline{60.8}              \\ \midrule
    \multirow{5}{*}{{\begin{tabular}[c]{@{}c@{}}General \\ \&\\ Medical\\ LLMs\end{tabular}}}            & Mistral(7B)                      & Mistral2023                         & 63.4($\pm$0.4)                                                                          & 47.7($\pm$0.7)              & 64.4($\pm$0.1)              & 58.5                          \\
                                                                                                         & MedAlpaca(7B)                    & BHT2023                             & 60.0($\pm$0.4)                                                                          & 40.1($\pm$0.1)              & 49.3($\pm$3.4)              & 49.8                          \\
                                                                                                         & PMC-LLaMA(7B)                    & SJTU2024                            & 20.7($\pm$1.1)                                                                          & 24.7($\pm$0.4)              & 34.6($\pm$1.7)              & 26.7                          \\
                                                                                                         & LLaMA3(8B)                       & Meta2024                            & 63.4($\pm$0.5)                                                                          & {56.6($\pm$0.4)}            & 65.4($\pm$0.6)              & 61.8                          \\
                                                                                                         & \textbf{LLaMA3.1(8B)[baseline]}  & Meta2024                            & \underline{67.7($\pm$0.7)}                                                              & \underline{56.3($\pm$0.6)}  & \underline{68.7($\pm$0.6)}  & \underline{64.2}              \\ \midrule
    \multirow{5}{*}{{\begin{tabular}[c]{@{}c@{}}General \\ \&\\ Medical\\ LLMs \\with COT\end{tabular}}} & COT-Mistral(7B)                  & Mistral2023                         & 63.4($\pm$0.3)                                                                          & 47.4($\pm$0.2)              & 65.1($\pm$0.2)              & 58.6                          \\
                                                                                                         & COT-MedAlpaca(7B)                & BHT2023                             & 60.3($\pm$0.4)                                                                          & 39.9($\pm$0.3)              & 48.5($\pm$2.5)              & 49.6                          \\
                                                                                                         & COT-PMC-LLaMA(7B)                & SJTU2024                            & 20.4($\pm$0.8)                                                                          & 20.8($\pm$0.2)              & 20.8($\pm$0.6)              & 20.7                          \\
                                                                                                         & COT-LLaMA3(8B)                   & Meta2024                            & 65.1($\pm$0.5)                                                                          & \underline{55.2($\pm$0.3)}  & 64.2($\pm$0.5)              & 61.5                          \\
                                                                                                         & COT-LLaMA3.1(8B)                 & Meta2024                            & \underline{68.1($\pm$0.5)}                                                              & 54.9($\pm$0.3)              & \underline{70.6($\pm$0.5)}  & \underline{64.5 }             \\\midrule
    \multirow{4}{*}{{\begin{tabular}[c]{@{}c@{}}RAG \\ Based\\ Methods\end{tabular}}}                    & Self-RAG (7B)                    & ICLR2024                            & 32.2 ($\pm$1.9)                                                                         & 38.0 ($\pm$2.8)             & 59.4 ($\pm$1.2)             & 43.2                          \\
                                                                                                         & Self-RAG (13B)                   & ICLR2024                            & 50.2 ($\pm$0.4)                                                                         & 40.8 ($\pm$2.0)             & 64.6 ($\pm$5.0)             & 51.9                          \\
                                                                                                         & KG-Rank (13B)                    & ACL-w2024                           & 45.2 ($\pm$0.5)                                                                         & 36.2 ($\pm$1.1)             & 50.3 ($\pm$1.5)             & 43.9                          \\
                                                                                                         & MedRAG (70B)                     & Oxon2024                            & \underline{57.9 ($\pm$1.5)}                                                             & \underline{48.7 ($\pm$1.4)} & \underline{71.9 ($\pm$1.8)} & \underline{59.5}              \\
    \midrule
    \multirow{1}{*}{{\begin{tabular}[c]{@{}c@{}}Logic Based \end{tabular}}}                              & MedLA + LLaMA3.1(8B)             & Ours                                & \textbf{70.7($\pm$0.1)}                                                                 & \textbf{62.6($\pm$0.1)}     & \textbf{76.5($\pm$0.1)}     & \textbf{69.9($\uparrow$5.7)}  \\
    \bottomrule
  \end{tabular}
   \caption{{The performance of our proposed MedLA model on Multi-choice medical QA benchmarks \citep{xiong2024medrag}.} The table includes the accuracy (Acc) along with the standard deviation ($\pm$std) for each metric. The results demonstrate the effectiveness of MedLA in addressing complex medical reasoning tasks across different datasets. }\label{tab_results_medicalqa1}
\end{table*}

%% file: tab_expert.tex
\begin{table}[t]
\centering

\begin{tabular}{lll r}
\toprule
\textbf{Model}& \textbf{Method} & \textbf{Number} & \textbf{Score} \\
\midrule
deepseek-r1& MedLA    & 60 & 36.0($\pm$4.3)\\
      & baseline & 60 & 21.3($\pm$4.9) \\
deepseek-v3& MedLA    & 60 & 25.6($\pm$3.1)\\
      & baseline & 60 & 15.0($\pm$0.0)\\
\bottomrule
\end{tabular}
\caption{Performance comparison of \textbf{MedLA} and \textbf{baseline} on DeepSeek-based reasoning evaluated on the MedXpertQA benchmark\citep{zuo2025medxpertqa}.}\label{tab:deepseek_medical_reasoning}
\end{table}

%% file: tab_smaltab.tex
\begin{table}[t]
    \centering
    \footnotesize
    \begin{tabular}{p{2.2cm}p{1.5cm}p{1.5cm}p{1.5cm}}
        \toprule
        \multirow{2}{*}{Variant} & \textsc{MedQA-US}     & \multicolumn{2}{c}{\textsc{MedDDx}}                         \\
        \cmidrule(lr){3-4}
                                 & Acc$\pm$std           & Basic                               & Expert                \\ \midrule
        \textsc{MedLA} (full)    & \textbf{62.6}$\pm$0.1 & \textbf{48.2}$\pm$2.1               & \textbf{41.7}$\pm$0.8 \\[2pt]
        \;-Revision loop         & 58.4$\pm$0.3          & 44.2$\pm$1.9                        & 38.6$\pm$1.0          \\
        \;-Credibility           & 57.3$\pm$0.4          & 41.8$\pm$1.7                        & 37.2$\pm$1.3          \\
        \;-LogicTree (CoTOnly)  & 56.1$\pm$0.4          & 38.7$\pm$1.5                        & 34.9$\pm$1.2          \\ \midrule
        MV     & 54.8$\pm$0.4          & 37.5$\pm$0.9                        & 30.2$\pm$0.8          \\
        \bottomrule
    \end{tabular}
    \caption{{Step-wise ablation of \textsc{MedLA}.} The last row Majority Voting~(\textsc{MV}) is a naïve majority-vote ensemble that keeps only the common backbone (\textsc{LLaMA 3.1-8B}) and the same agent prompts. All numbers are three-seed means $\pm$\,std.}
    \label{tab_ablation}
\end{table}
\begin{table}[t]
    \centering
    \footnotesize
    \begin{tabular}{lrrrrr}
        \toprule
        \multirow{2}{*}{\textbf{Method}} & \multicolumn{4}{c}{\textbf{Component time (sec.)}} & \multirow{2}{*}{\textbf{Total}}                                              \\
        \cmidrule(lr){2-5}
                                         & \textbf{FT}                                        & \textbf{RT}                     & \textbf{GBT} & \textbf{IFT(sec.)} &        \\ \midrule
        Majority Voting                  & --                                                 & --                              & --           & 1\,853             & 1\,853 \\
        MedAgents                        & --                                                 & --                              & --           & 2\,793             & 2\,793 \\
        KG-RAG                           & --                                                 & 603                             & --           & 1\,845             & 2\,448 \\
        KGAREVION                        & 10k+                                               & --                              & --           & 1\,821             & 10k+   \\
        \textbf{MedLA (ours)}            & --                                                 & --                              & --           & 3\,657             & 3\,657 \\ \bottomrule
    \end{tabular}
    \caption{{Time Consumption Analysis.} Wall-clock~(sec.) latency on \textsc{BioASQ-Y/N}. FT = extra fine-tuning, RT = retrieval, GBT = logic-graph build (incl. revision loop), IFT = pure LLM inference. `--' indicates no time cost.}
    \label{tab_time}
\end{table}

%% file: fig_small.tex


\begin{figure}[t]
    \centering
    \includegraphics[width=0.99\columnwidth]{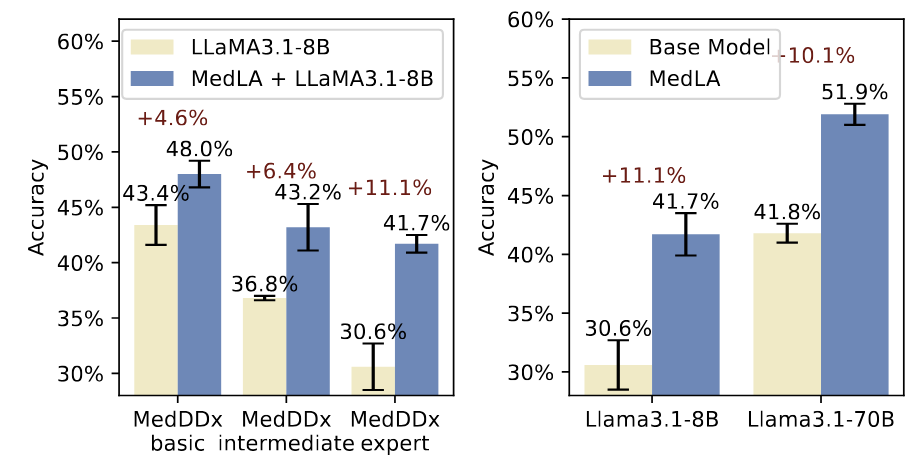}
    \caption{Performance comparison of MedLA with LLaMA3.1-8B at different levels of difficulty on the MedDDx benchmark. Error bars represent SD.}
    \label{fig_MedDDx}
\end{figure}

%% file: app.tex
\section{Appendix: Details of Related Work} \label{Appendix_relatedwork}
\textbf{Large Language Models in Medical Reasoning.}
Large language models (LLMs) have demonstrated strong capabilities in general-purpose natural language understanding and reasoning tasks~\citep{brown2020gpt3}. In the medical domain, LLMs have been widely applied to diagnosis assistance, clinical report generation, and medical question answering. Models such as GPT-3 and its successors have set new benchmarks in language modeling, yet they often lack the domain-specific knowledge and structured reasoning abilities required to handle semantically complex clinical scenarios~\citep{liu_toolnet_2024}. To address this, researchers have introduced Chain-of-Thought (CoT) prompting~\citep{wei2022cot}, which decomposes complex problems into interpretable reasoning steps. While effective for generic tasks, CoT struggles with the multi-modal, context-dependent nature of medical problems~\citep{liu2023llavamed}. Med-Flamingo extended CoT to a multi-modal setting, combining textual and visual inputs to explore richer clinical reasoning pathways~\citep{moor2023medflamingo}. However, despite such advances, most LLMs remain constrained by static reasoning structures and closed pretraining knowledge, underscoring the need for adaptive, context-aware mechanisms~\citep{sepehri_mediconfusion_2024, zhou_reliable_2024}.

\textbf{Multi-Agent Systems for Collaborative Problem Solving.}
Multi-agent systems (MAS) have gained traction as an effective paradigm for solving complex tasks, especially those requiring multidisciplinary expertise and collaborative reasoning~\citep{kim2024mdagents}. In the medical domain, these systems typically assign fixed roles to agents, such as radiologist, pathologist, or general practitioner, to simulate real-world clinical collaboration. For example, MedAgents adopts static role-based assignments to facilitate zero-shot medical question answering by leveraging pretrained LLMs to emulate specialist reasoning~\citep{tang2024medagents}. Similarly, MDAgents introduces a hierarchical coordination framework to adaptively structure collaboration based on task complexity, significantly improving both flexibility and diagnostic accuracy~\citep{kim2024mdagents}. Other approaches, such as GopherCite~\citep{rae2021gophercite}, integrate retrieval-augmented mechanisms to inject domain-specific knowledge into the generation process. Nevertheless, most existing systems rely heavily on predefined role structures, limiting their capacity to generalize across novel tasks or dynamically evolving medical scenarios~\citep{guo2024large}.

\textbf{Dynamic Multi-Agent Collaboration.}
To overcome the limitations of static role assignment, researchers have explored dynamic agent collaboration mechanisms. MMedAgent introduces a multi-modal tool interface that enables agents to dynamically select the most appropriate tools based on task requirements, significantly improving flexibility and precision in multi-modal medical tasks~\citep{li2024mmedagent}. The system integrates modules such as segmentation, classification, and report generation, supporting a wide range of end-to-end workflows. RETRO similarly enhances factuality by combining LLMs with large-scale retrieval systems~\citep{borgeaud2022retro}. While these systems offer improved task-level adaptability, their focus remains on tool orchestration rather than deeper inter-agent collaboration or logical reasoning. For example, Med-Flamingo and LLaVA-Med demonstrate effective multi-modal integration but lack structured support for multi-agent reasoning across logical layers.

\textbf{Logic-Driven Frameworks in Medical AI.}
Recent studies emphasize the critical role of logical reasoning in enhancing the problem-solving capabilities of medical AI systems~\citep{tang2024medagents}. Logic-driven frameworks decompose tasks into interpretable modules, such as causal inference, diagnosis validation, and treatment planning, enabling greater transparency and robustness. MedAgents leverages hierarchical reasoning pipelines to emulate expert workflows, while MDAgents dynamically structures collaboration based on task complexity to improve performance on complex cases~\citep{kim2024mdagents}. Other approaches, such as Logic-Guided GPT~\citep{ye2022logicguided}, explicitly embed symbolic logic into LLMs, improving controllability and reasoning depth in structured domains. Despite these efforts, most existing frameworks remain bounded by static task definitions and lack mechanisms for deep agent-to-agent reasoning. The integration of logic-driven reasoning with dynamic multi-agent collaboration remains an open challenge, especially in addressing the multi-layered logic and multi-perspective inference required for real-world clinical reasoning~\citep{wu2024medical, su2024knowledge}.

\section{Appendix: Details of PROMPTS} \label{Appendix_PROMPTS}
\label{Appendix_prompts}

In this section, we provide the detailed prompt templates used for each agent in the MedLA framework. The prompts are divided into two parts: system instruction, which defines the task and formatting conventions, and user instruction, which provides a case study and describes the specifics of the task. For all experiments, prompts are written in English, and the user prompts are formatted to follow a consistent pattern: \texttt{[Example]} + \texttt{[Context]} + \texttt{[Expected Output Format]}. All prompt templates are manually curated and kept fixed during testing. Below, we list the prompt classes and examples. And depending on the task, Agent's chat logs are selectively saved.

\subsection{D-Agent: Decompose Agent Prompt}

\textbf{System Prompt:}
\begin{quote}
\small

You are an agent. Help me eliminate the least likely option. Make sure to follow the format below: \\
$<$Eliminate$>$Answer: {option}$<$/Eliminate$>$ \\
Reason: *** \\
(don't forget the $<$Eliminate$>$ $<$/Eliminate$>$ tags.)
\end{quote}

\textbf{Purpose:} Remove the most unlikely answer from multiple choices based on logic down to the last option. Rebuttal with other agents if desired.

\textbf{User prompt template:}
\begin{quote}
\small

Based on the following examples, return the answer to the medical question. Examples are as follows: \\
\texttt{[Examples]} \\
Question: \texttt{[question]}\\ 
Option: \texttt{[option]}\\ 
Help me eliminate the least likely option. Then explain in one sentence the key reason for such a judgment. Please follow the structured reasoning steps below and provide the final answer clearly: \\
Step 1: Information Extraction  \\
List clearly each important piece of information provided in the question:\\
- Information 1: ... \\
- Information 2: ... \\
- Information 3: ... \\
(Continue as needed) \\
Step 2: Background Knowledge  \\
Clearly state relevant background knowledge or general principles necessary for solving this problem: \\
- Knowledge 1: ... \\
- Knowledge 2: ... \\
- Knowledge 3: ... \\
(Continue as needed) \\
Step 3: Reasoning Process (Subject-Predicate-Object format)   \\
Perform logical reasoning step-by-step, explicitly indicating each reasoning step as a Subject-Predicate-Object triple, along with a brief explanation of the logical relationship: \\
Reasoning Step 1: \\
- Subject: ... \\
- Predicate (relation): ... \\
- Object: ... \\
- Explanation: ... \\
Reasoning Step 2: \\
- Subject: ... \\
- Predicate (relation): ... \\
- Object: ... \\
- Explanation: ... \\
Reasoning Step 3: \\
- Subject: ... \\
- Predicate (relation): ... \\
- Object: ... \\
- Explanation: ... \\
(Continue additional reasoning steps as necessary) \\
- Object: The entity or concept that receives the action or judgment. \\
---- \\
Follow the format below:  \\
$<$Eliminate$>$Answer: \texttt{[option\_str]}$<$/Eliminate$>$  \\
Reason: ***  \\
(don't forget the $<$Eliminate$>$ $<$/Eliminate$>$ tags.) 
\end{quote}

\textbf{Purpose:} Rebuttal with other agents if desired.

\textbf{User prompt template:}
\begin{quote}
\small
Question:\texttt{[question]}  \\
Your opinion is: \texttt{[answer]}. Your reasoning is: \texttt{[reason]}  \\
Please communicate with multiple other logical perspectives.  \\
\texttt{[opinions]}  \\
First answer, which should be the correct option.  \\
Then, please summarize each person's logic, and if you find them to be factually incorrect, and logically incorrect, please list them below and give reasons.  \\
-----  \\
Follow the format below:   \\
$<$Answer$>$Answer: \texttt{[options]}$<$/Answer$>$.  \\
(don't forget the $<$Answer$>$ $<$/Answer$>$ tags.)  \\
\end{quote}

\subsection{M-Agent: Medical Agents}

\textbf{System Prompt:}
\begin{quote}
\small
You are a professional medical logic checker. Your task is to analyze a given medical reasoning passage and identify each reasoning step, its logical structure, and potential issues. In addition to a written evaluation, you must output a TSV-style table summarizing each reasoning step. Make sure your TSV-style output is accurate, logically sound, and grounded in standard medical knowledge. \\
Just return this TSV form. \\

Example of the table format: \\
Step Subject Object Logical Relationship Reasoning Text Credibility Error (Yes/No) Error Type Suggested Correction   \\
Fever Bacterial Infection Symptom-to-cause The presence of fever indicates a bacterial infection Weak Yes Factual Clarify that fever is non-specific and may be caused by viral or bacterial infection  \\
\end{quote}

\textbf{Purpose:} Go through and summarize each of the L-Agent's gave and organize them in TSV format.

\textbf{User prompt template:}
\begin{quote}
\small
reasoning passage:\texttt{[reason]} \\

Please perform the following tasks: \\
1. [Step-by-step Reasoning Reconstruction]   \\
Break down the entire reasoning process step-by-step. Identify: \\
- The subject (who or what the step is about) \\
- The object (the outcome, result, or related concept) \\
- The logical relationship (e.g., cause-effect, association, inference) \\
- The logical strength (Strong / Moderate / Weak) \\
- Whether the step is flawed (Yes / No) \\
- If flawed, specify the error type (Factual / Logical / Conceptual) \\

2. [TSV Table Output]   \\
Provide your analysis in TSV table format with the following columns: \\
Step, Subject, Object, Logical Relationship, Reasoning Text, Credibility, Error (Yes/No), Error Type, Suggested Correction \\

3. [Optional Full Report]   \\
After the TSV table, you may include a full written report including: \\
- Summary of reasoning \\
- Overall assessment of reasoning strength \\
- Suggestions for correction \\

Example of the table format: \\
Step Subject Object Logical Relationship Reasoning Text Credibility Error (Yes/No) Error Type Suggested Correction   \\
Fever Bacterial Infection Symptom-to-cause The presence of fever indicates a bacterial infection Weak Yes Factual Clarify that fever is non-specific and may be caused by viral or bacterial infection   \\
... \\

Make sure your TSV-style output is accurate, logically sound, and grounded in standard medical knowledge. \\
Just return this TSV form.
\end{quote}

\subsection{C-Agent: Credibility Agent Prompt}

\textbf{System Prompt:}
\begin{quote}
\small
You are the Credibility Agent responsible for assessing the trustworthiness of each inference step in a medical logical tree.\\
Your task is to review each syllogism node—its major premise, minor premise, and conclusion—and assign one of three credibility levels: {High}, {Medium}, or {Low}.\\
Follow the output format exactly as a TSV table with columns: \\
\texttt{Index \;\; MajorPremise \;\; MinorPremise \;\; Conclusion \;\; Credibility}\\
Do not include any extra commentary.
\end{quote}

\textbf{Purpose:}  
Evaluate the logical consistency, factual correctness, and relevance of each node in the agent-generated tree, so that downstream discussion can focus on low-credibility steps.

\textbf{User Prompt Template:}
\begin{quote}
\small
Here is the logical tree from Medical Agent \texttt{M\textsuperscript{(j)}}. Each row is a syllogism node:

\medskip
\begin{tabular}{ll}
\textbf{Index} & \textbf{Syllogism} \\
\hline
1 & (All X cause Y)  (Patient has X) $\to$ (Patient may have Y) \\
2 & (Rule A)  (Fact B) $\to$ (Conclusion C) \\
… & … \\
\end{tabular}

\medskip
For each node, decide:
\begin{itemize}
  \item Is the major premise correct and relevant?
  \item Is the minor premise factual and applicable?
  \item Does the conclusion follow logically?
\end{itemize}

Assign {High}, {Medium}, or {Low} credibility to each node and return a TSV table:

\medskip
\texttt{Index \;\; MajorPremise \;\; MinorPremise \;\; Conclusion \;\; Credibility}

\medskip
Example row:
\texttt{1 \;\; "All X cause Y" \;\; "Patient has X" \;\; "Patient may have Y" \;\; High}
\end{quote}

\subsection{P-Agent: Premise Agent Prompt}

\textbf{System Prompt:}
\begin{quote}
\small
You are the Premise Agent responsible for extracting the foundational facts and rules from a clinical question.\\
Your output must include two labeled lists: \texttt{$<$MajorPremises$>$} and \texttt{$<$MinorPremises$>$}.\\
Each list should contain bullet-style items, one premise per line.\\
Do not include any additional text or commentary.
\end{quote}

\textbf{Purpose:}  
Seed the logical reasoning pipeline by isolating (1) general medical rules (major premises) and (2) patient-specific facts (minor premises) from the question text.

\textbf{User Prompt Template:}
\begin{quote}
\small
Below is a medical question. Extract:
\begin{itemize}
  \item All relevant general medical rules that could apply (\emph{major premises}).
  \item All patient-specific facts stated in the question (\emph{minor premises}).
\end{itemize}

\medskip
\texttt{[Examples]}\\
Question: ``A 65-year-old man with hypertension presents with chest pain radiating to his left arm. What is the most likely diagnosis?''\\
\medskip
\texttt{[Context]}\\
Question: \texttt{[question text]}\\

\medskip
\texttt{[Expected Output Format]}\\
$<$MajorPremises$>$
\begin{itemize}
  \item “Hypertension increases risk of coronary artery disease.”
  \item “Chest pain radiating to left arm is a sign of myocardial ischemia.”
  \item ...
\end{itemize}
$<$/MajorPremises$>$

$<$MinorPremises$>$
\begin{itemize}
  \item “Patient is 65 years old.”
  \item “Patient has a history of hypertension.”
  \item “Patient’s chest pain radiates to the left arm.”
  \item ...
\end{itemize}
$<$/MinorPremises$>$
\end{quote}

\section{Appendix: Examples of Logical Tree} \label{Appendix_tree}

\subsection{Example Question}
The following example illustrates the end-to-end process by which a single agent constructs a logical tree.

\begin{itemize}
  \item \textbf{Question}: Which medications are designed to influence genes or proteins associated with the enzymatic activity of dITP diphosphatase?
  \item \textbf{Options}:
    A. Sodium citrate  
    B. Citric acid  
    C. Calcium citrate  
    D. Ascorbic acid  
\end{itemize}

\subsection{Generated Logical Tree by One Agent}
Figures~\ref{fig_logic_tree_1} (Part 1) and~\ref{fig_logic_tree_2} (Part 2) show the logical tree produced by a single M-Agent for the example question, alongside its text-based reasoning panel. Below is the complete English text that appears in the left-hand panel.

\paragraph{Step 1 - Background Knowledge Retrieval}
\begin{enumerate}
  \item \textbf{Knowledge 1}: Drugs intended to affect dITP diphosphatase typically bind directly to the enzyme or regulate its gene/protein expression to alter its catalytic activity.
  \item \textbf{Knowledge 2}: Buffering or chelating salts (e.g., sodium citrate) are formulated for pH control or anticoagulation and are not engineered for selective enzyme modulation.
  \item \textbf{Knowledge 3}: Vitamins acting as cofactors (e.g., ascorbic acid) can participate in redox or epigenetic pathways, thereby modulating gene expression.
  \item \textbf{Knowledge 4}: Compounds that raise intracellular Ca\textsuperscript{2+} (e.g., calcium citrate) activate Ca\textsuperscript{2+}-dependent transcription factors, indirectly influencing enzyme expression.
  \item \textbf{Knowledge 5}: TCA-cycle metabolites (e.g., citric acid) may feedback-regulate metabolic-gene networks, providing a theoretical link to nucleotide-related enzymes.
\end{enumerate}

\paragraph{Step 2 - Information Extraction}
\begin{enumerate}
  \item \textbf{Information 1}: The task is to eliminate the medication least likely to modulate dITP diphosphatase activity.
  \item \textbf{Information 2}: The options are A = Sodium citrate, B = Citric acid, C = Calcium citrate, D = Ascorbic acid.
  \item \textbf{Information 3}: dITP diphosphatase belongs to the nucleotide-sanitation pathway; modulators usually derive from nucleotide or gene-regulatory pharmacology (relates to Knowledge 1).
\end{enumerate}

\subsection{Step 3 - Reasoning Process (Subject-Predicate-Object and Explanation)}

\begin{tabular}{c|l|p{3.5cm}|p{3cm}|p{3cm}}
\toprule
\textbf{Step} & \textbf{Subject}    & \textbf{Predicate (relation)}                                           & \textbf{Object (conclusion)}              & \textbf{Explanation (Knowledge $\times$ Information)} \\ \midrule
1           & Sodium citrate     & Lacks any engineered gene- or enzyme-modulatory design (Knowledge 2)     & $\Rightarrow$ Eliminate A                 & K2 + I1  \\
2           & Ascorbic acid      & Acts as a redox cofactor and can alter gene expression epigenetically (Knowledge 3) & $\Rightarrow$ Keep B                     & K3 + I3  \\
3           & Calcium citrate    & Raises intracellular Ca\textsuperscript{2+}, activating Ca\textsuperscript{2+}-dependent transcription factors (Knowledge 4) & $\Rightarrow$ Keep C    & K4 + I3  \\
4           & Citric acid        & Feeds back on metabolic-gene networks (Knowledge 5)                      & $\Rightarrow$ Keep D                      & K5 + I3  \\ \bottomrule
\end{tabular}

\paragraph{Final Conclusion}  
\texttt{$<$Eliminate$>$Answer: A$<$/Eliminate$>$}  
\quad Reason: Sodium citrate is merely a buffering/anticoagulant salt and is not designed to modulate nucleotide-metabolizing enzymes such as dITP diphosphatase.

\begin{figure}[!p]
    \centering
    \includegraphics[width=0.95\textwidth]{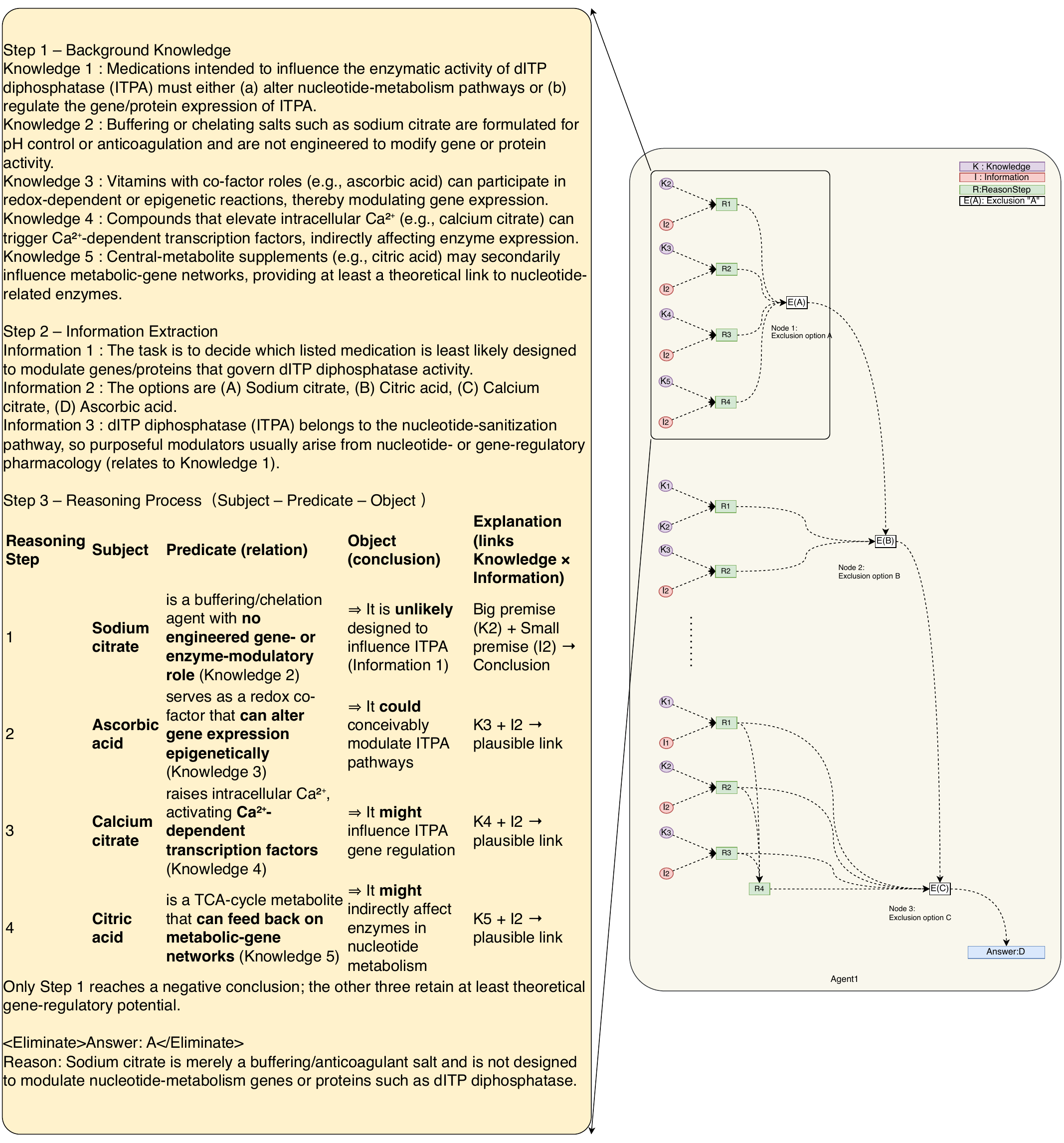}
    \caption{\textbf{The first part of the logical tree generated by Agent1 for the example question.} The tree is derived from the question and retrieved documents, where nodes represent reasoning steps.}
    \label{fig_logic_tree_1}
\end{figure}

\begin{figure}[!p]
    \centering
    \includegraphics[width=0.95\textwidth]{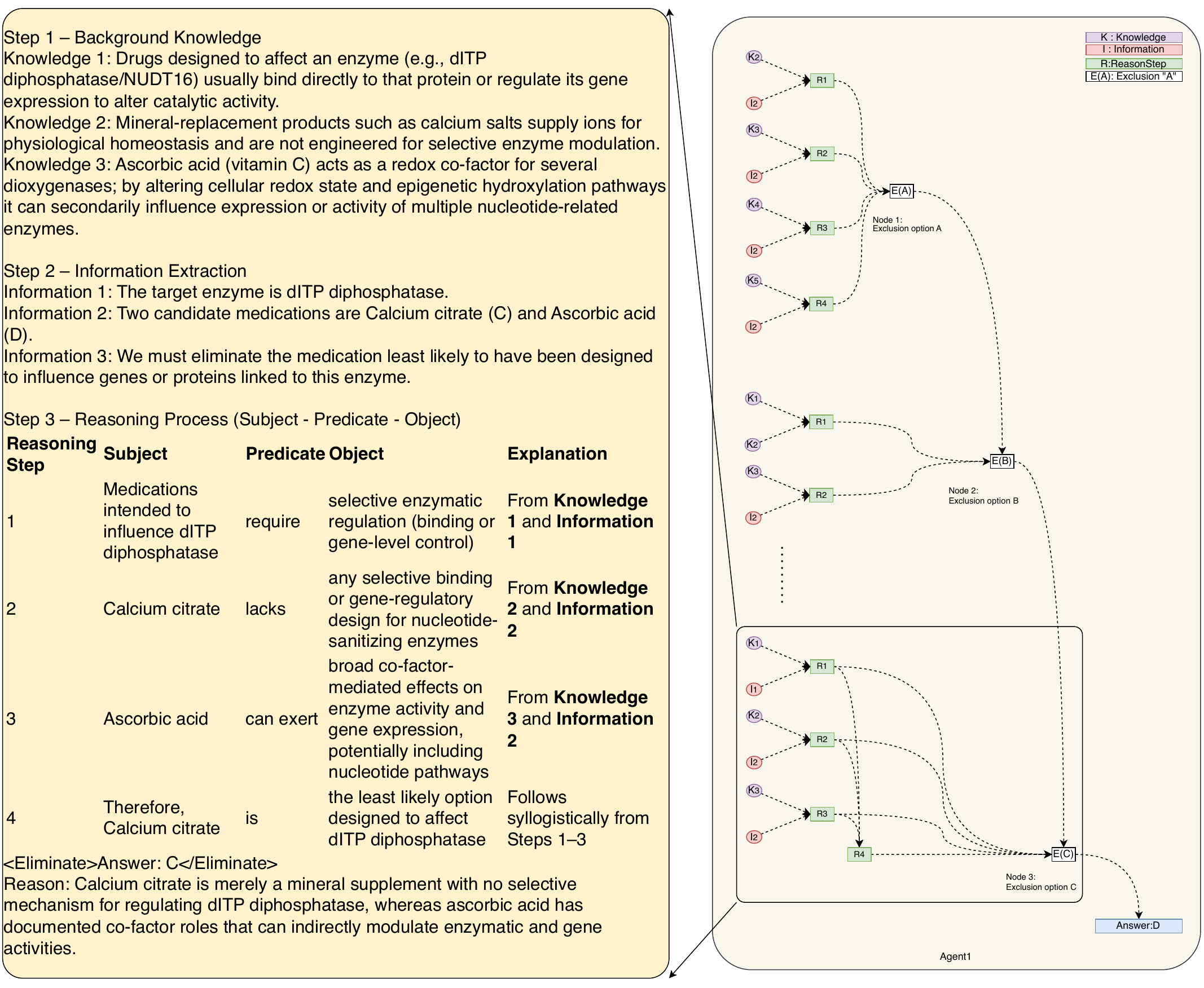}
    \caption{\textbf{The second part of the logical tree generated by Agent1 for the example question.} This figure continues the visualization of the reasoning process shown previously.}
    \label{fig_logic_tree_2}
\end{figure}

\vspace{0.5em}
The text above corresponds exactly to the nodes and edges in Figures~\ref{fig_logic_tree_1} and~\ref{fig_logic_tree_2}. It shows how the agent leverages background knowledge and question-specific information to build and verify syllogistic reasoning steps, culminating in the elimination of option A.

\paragraph{Legend}
\begin{itemize}
  \item \textbf{K}: Knowledge node  
  \item \textbf{I}: Information node  
  \item \textbf{R}: Reasoning step node  
  \item \textbf{E(option)}: Elimination/Selection node  
\end{itemize}

Figures~\ref{fig_logic_tree_1} and~\ref{fig_logic_tree_2} together illustrate the agent’s three-stage output: the left panel for textual reasoning and the right panel for the corresponding visualized logical tree.

\section{Appendix: Details of Dataset \& Benchmarks} \label{Appendix_datset}
\begin{table}[t]
    \centering

    \begin{tabular}{l|l|l}
        \toprule
         Datasets            & Size  & QA            \\ \midrule
         MMLU-Med            & 1,089 & A/B/C/D       \\
         MedQA-US            & 1,273 & A/B/C/D       \\
         BioASQ-Y/N          & 618   & Yes/No        \\
         MedDDx-Basic        & 245   & A/B/C/D       \\
         MedDDx-Intermediate & 1,041 & A/B/C/D       \\
         MedDDx-Expert       & 483   & A/B/C/D       \\
         Medxpert(random samples)&60 &A/B/C/D/...\\
         
        \bottomrule
    \end{tabular}
    \caption{Summary of datasets used in our experiments. The table includes the dataset names, sizes, question-answering (QA) formats, and whether they are open-ended reasoning (OR) tasks.}
    \label{tab_supp_dataset}
\end{table}

\subsection{Details of the MedDDx}

MedDDx is a newly constructed dataset designed to test model performance on semantically complex answers. The motivation behind creating this dataset is twofold:

While LLMs can perform QA tasks, they often rely heavily on semantic dependencies, making it difficult for them to identify the correct answer among semantically similar answer candidates.

In real-world medical scenarios, researchers often focus on identifying subtle differences between similar molecules, particularly in treatment or diagnostic settings. For instance, proteins may share similar names but have significantly different structures and functions, making it crucial to distinguish these differences to be able to provide accurate answers (see Table \ref{tab:qa-difficulty} for an example).

\begin{table}[t]
    \centering
    
    \begin{tabular}{@{}p{2.5cm}>{\raggedright\arraybackslash}p{6cm}@{}}
        \toprule
        \textbf{Dataset Name} & \textbf{Sample} \\
        \midrule
        Basic & Can you recommend medications effective against peptic ulcer disease that also suppress \textit{Helicobacter pylori} in the stomach? \\
              & A: Rebamipide \quad B: Ecabet \quad C: Bendazac \quad D: Nepafenac \\
        \midrule
        Intermediate &  Can you recommend medications that treat both eosinophilic pneumonia and a parasitic worm infection? \\
                     & A: Thiabendazole \quad B: Albendazole \quad C: Diethylcarbamazine \quad D: Triclabendazole \\
        \midrule
        Expert & Which genes or proteins are expressed exclusively in the pericardium and not in either the dorsal or ventral regions of the thalamus? \\
               & A: ADH1A \quad B: ADH1C \quad C: ADH4 \quad D: ADH1B \\
        \bottomrule
    \end{tabular}
    \caption{Examples of medical multiple-choice questions at different difficulty levels.}\label{tab:qa-difficulty}
\end{table}

Because of these reasons, \citep{su2025kgarevion} constructs MedDDx, a multi-choice medical QA dataset that focuses on answering semantically complex multi-choice QA. These questions are sourced from STaRK-Prime (Wu et al., 2024c), which provides both the questions and their corresponding answers. \citep{su2025kgarevion} extract questions with a single correct answer from the STaRK-Prime testing set and transform them into the multiple-choice format. To generate three strong alternative answer candidates, \citep{su2025kgarevion} uses semantic similarity to increase the difficulty, selecting the top three entities that have the highest semantic similarity as the correct answer. The semantic embeddings used for this process are derived from the text-embedding-ada-002 model from OpenAI.  The semantic similarity is calculated using cosine similarity. \citep{su2025kgarevion} also computes the standard deviation of semantic similarity between the correct answer and the other three candidates. The density distribution of these values is shown in Fig. 8. Based on this distribution, \citep{su2025kgarevion} divides the queries into three complexity groups using quantile analysis: MedDDx-Expert (0-0.02), MedDDx-Intermediate (0.02-0.04), and MedDDx-Basic ($>0.04$).

\subsection{Details of the multi-choice medical QA benchmarks}

In this work, we use four well-known multi-choice medical QA datasets to evaluate the model performance, including two medical examination QA datasets (MMLU-Med, MedQA-US) and two biomedical research QA datasets (PubMedQA*, BioASQ-Y/N). These datasets are derived from (Xiong et al., 2024a). The samples in these datasets are shown in Table \ref{tab:biomed-qa-examples}.

\subsection{Details of the MedXpertQA }
MedXpertQA is a newly introduced, highly challenging, and comprehensive benchmark designed to evaluate expert-level medical knowledge and advanced reasoning. Recognizing that medicine provides a rich setting for assessing real-world decision-making beyond mathematics and code, this benchmark was developed to push the boundaries of clinical evaluation. It comprises 4,460 questions spanning 17 medical specialties and 11 body systems.

To address the insufficient difficulty of existing benchmarks like MedQA, MedXpertQA was constructed through rigorous filtering and augmentation, incorporating questions from specialty board exams to enhance clinical relevance and comprehensiveness. The development process included data synthesis to mitigate data leakage risks and multiple rounds of expert reviews to ensure accuracy and reliability.

\begin{table}[t]
    \centering
  
    \begin{tabular}{@{}p{1cm}>{\raggedright\arraybackslash}p{6cm}@{}}
        \toprule
        \textbf{Dataset Name} & \textbf{Sample} \\
        \midrule
        \multirow{2}{=}{MMLU-Med} & Which of the following best describes the structure that collects urine in the body? \\
                                  & A: Bladder \quad B: Kidney \quad C: Ureter \quad D: Urethra \\
        \midrule
        \multirow{2}{=}{MedQA-US} & A microbiologist is studying the emergence of a virulent strain of the virus. After a detailed study of the virus and its life cycle, he proposes a theory: Initially, a host cell is co-infected with 2 viruses from the same virus family... (truncated for brevity). \\
                                  & A: Epstein-Barr virus \quad B: Human immunodeficiency virus \quad C: Rotavirus \quad D: Vaccinia virus \\
        \midrule
        \multirow{2}{=}{BioASQ-Y/N} & Can losartan reduce brain atrophy in Alzheimer's disease? \\
                                   & A: yes \quad B: no \\
        \bottomrule
    \end{tabular}
      \caption{Examples of questions from multi-choice medical QA benchmarks.}\label{tab:biomed-qa-examples}
\end{table}
    
    \textbf{Datasets\& Benchmarks.}  
We evaluate on three complementary benchmarks.
\noindent\textbf{(i) {MedRAG-QA}} \citep{xiong2024medrag}.  
This suite aggregates three established multiple-choice datasets—{MMLU-Med}, {MedQA-US}, and {BioASQ-Y/N}.  
It covers factual recall, guideline interpretation, and clinical decision-making ( details in Table~\ref{tab_supp_dataset}).  
The benchmark serves as our {general-purpose} test bed.
\noindent\textbf{(ii) {MedDDx}.}  
To stress differential-diagnosis reasoning, we construct {MedDDx} following the protocol of \citep{su2025kgarevion}.  
From the {STaRK-Prime} knowledge base \citep{wu2024stark} we extract case-answer pairs and attach three semantically closest but incorrect disorders as distractors.  
Difficulty is stratified by the standard deviation of the answer-distractor similarity into {Basic}, {Intermediate}, and {Expert} tiers, forcing increasingly fine-grained causal discrimination.
\noindent\textbf{(iii) {MedXpertQA}\citep{zuo2025medxpertqa}} 
To assess performance in the most demanding clinical scenarios that require expert-level knowledge, we utilize MedXpertQA. This benchmark is comprised of challenging questions derived from medical board certification exams and specialist case studies. It is specifically designed to test advanced clinical reasoning, requiring the synthesis of complex patient information and multi-step diagnostic logic, serving as our primary benchmark for evaluating advanced clinical intelligence.

\section{Appendix: Specifications of Theoretical Properties }
\label{Appendix_theory}
This section provides the rigorous theoretical specifications for the convergence and stability of the MedLA framework. The framework is designed to satisfy two key properties: (1) monotonic reduction of internal disagreement among agents and (2) guaranteed convergence to a stable state in a finite number of rounds.

\subsection*{Core Model Definitions and Assumptions}

The theoretical model is based on the following components and operational assumptions:

\begin{itemize}
    \item \textbf{Logic ($T_t^{(j)}$):} A representation of the reasoning process for the $j$-th agent at round $t$. We assume each logic belongs to a finite set of all possible unique logics, $\mathcal{T}$.

    \item \textbf{Conclusion Variable ($C_t^{(j)}$):} A scalar value representing the conclusion of the $j$-th agent. The conclusion is a deterministic function of the agent's logic, i.e., $C_t^{(j)} = f(T_t^{(j)})$, where $f: \mathcal{T} \to \mathbb{R}$.

    \item \textbf{System State ($L_t$):} The complete state of the system at round $t$, defined by the set of all agents' logics: $L_t = \{T_t^{(1)}, T_t^{(2)}, \dots, T_t^{(N_M)}\}$. The system state space, $\mathcal{L} = \mathcal{T}^{N_M}$, is finite.

    \item \textbf{Agent Consensus ($\bar{C}_t$):} The arithmetic mean of all agent conclusions:
    $$ \bar{C}_t = \frac{1}{N_M} \sum_{j=1}^{N_M} C_t^{(j)} $$

    \item \textbf{Internal Variance ($S_t^2$):} A measure of inter-agent disagreement, calculated as the sample variance of their conclusions:
    $$ S_t^2 = \frac{1}{N_M-1} \sum_{j=1}^{N_M} (C_t^{(j)} - \bar{C}_t)^2 $$

    \item \textbf{Correction Mechanism:} An agent updates its conclusion by adjusting its logic. The resulting numerical change is governed by the equation:
    $$ C_{t+1}^{(j)} = (1 - \alpha_{t,j})C_t^{(j)} + \alpha_{t,j}\bar{C}_t \label{eq:update} $$
    where the correction coefficient $\alpha_{t,j} \in [0,1]$.

    \item \textbf{Correction Condition:} A correction is performed ($T_t^{(j)} \to T_{t+1}^{(j)}$) only if an agent's logic $T_t^{(j)}$ is identified as suboptimal. Any non-trivial correction implies $\alpha_{t,j} > 0$. If no logic is updated, $\alpha_{t,j} = 0$ for all $j$.
    
    \item \textbf{Balanced Correction Assumption:} The aggregate effect of corrections is balanced such that the consensus center remains unchanged in a correction round. That is, if a correction occurs:
    $$ \sum_{j=1}^{N_M} \alpha_{t,j}(C_t^{(j)} - \bar{C}_t) = 0 $$
    This implies $\bar{C}_{t+1} = \bar{C}_t$.
\end{itemize}

\begin{remark}
The Balanced Correction Assumption formalizes the intuition that corrections from agents with conclusions above the mean are offset by corrections from agents with conclusions below the mean. This prevents the group consensus from drifting due to asymmetric updates.
\end{remark}

\begin{property}[Iterative Variance Reduction]
As long as at least one agent performs a correction (i.e., $\exists j$ such that $\alpha_{t,j} > 0$), the internal variance of the agent set strictly decreases: $S_{t+1}^2 < S_t^2$.
\end{property}

\begin{proof}
The internal variance at round $t+1$ is $S_{t+1}^2 = \frac{1}{N_M-1} \sum_{j=1}^{N_M} (C_{t+1}^{(j)} - \bar{C}_{t+1})^2$.

Under the Balanced Correction Assumption, the consensus center is invariant, $\bar{C}_{t+1} = \bar{C}_t$. We can analyze the deviation of an updated conclusion $C_{t+1}^{(j)}$ from this stable consensus center. Substituting the update rule from Eq. \eqref{eq:update}:
\begin{align}
    C_{t+1}^{(j)} - \bar{C}_t &= \left[ (1 - \alpha_{t,j})C_t^{(j)} + \alpha_{t,j}\bar{C}_t \right] - \bar{C}_t \nonumber \\
    &= (1 - \alpha_{t,j})C_t^{(j)} - (1 - \alpha_{t,j})\bar{C}_t \nonumber \\
    &= (1 - \alpha_{t,j})(C_t^{(j)} - \bar{C}_t) \label{eq:deviation}
\end{align}
Now, substituting this result into the definition of $S_{t+1}^2$:
\begin{align*}
    S_{t+1}^2 &= \frac{1}{N_M-1} \sum_{j=1}^{N_M} (C_{t+1}^{(j)} - \bar{C}_t)^2 \\
    &= \frac{1}{N_M-1} \sum_{j=1}^{N_M} \left[ (1 - \alpha_{t,j})(C_t^{(j)} - \bar{C}_t) \right]^2 \\
    &= \frac{1}{N_M-1} \sum_{j=1}^{N_M} (1 - \alpha_{t,j})^2 (C_t^{(j)} - \bar{C}_t)^2
\end{align*}
By the Correction Condition, there exists at least one agent $k$ for which $\alpha_{t,k} > 0$ and $C_t^{(k)} \neq \bar{C}_t$ (otherwise no correction would be needed). For this agent, $0 < \alpha_{t,k} \le 1$, which implies $0 \le (1 - \alpha_{t,k})^2 < 1$. For all other agents $j$, $\alpha_{t,j} \in [0, 1]$, so $(1 - \alpha_{t,j})^2 \le 1$.

Consequently, the weighted sum of squares must be strictly less than the original sum of squares:
\[
    \sum_{j=1}^{N_M} (1 - \alpha_{t,j})^2 (C_t^{(j)} - \bar{C}_t)^2 < \sum_{j=1}^{N_M} (C_t^{(j)} - \bar{C}_t)^2
\]
Dividing both sides by the positive constant $(N_M - 1)$ establishes the strict inequality:
\[
    S_{t+1}^2 < S_t^2
\]
This proves that any non-trivial correction round, under the assumption of balanced correction, monotonically and strictly reduces the internal variance.
\end{proof}

\begin{property}[Finite-Round Convergence]
The discussion process is guaranteed to converge to a stable fixed-point state within a finite number of rounds.
\end{property}

\begin{proof}
The proof proceeds by contradiction, based on a monotonic descent over a finite state space.

\begin{enumerate}
    \item \textbf{Finite State Space:} As defined previously, the system's complete state $L_t$ is an element of the state space $\mathcal{L} = \mathcal{T}^{N_M}$. Since the set of all possible logics $\mathcal{T}$ is finite and the number of agents $N_M$ is finite, the state space $\mathcal{L}$ is also finite.

    \item \textbf{Monotonic State-Value Function:} The internal variance $S_t^2$ is a function of the system state $L_t$, since each $C_t^{(j)}$ is determined by $T_t^{(j)} \in L_t$. Let this function be $V(L_t) = S_t^2$. As established in Property 1, any state transition $L_t \to L_{t+1}$ triggered by a correction implies a strict decrease in this value: $V(L_{t+1}) < V(L_t)$.

    \item \textbf{Termination and Preclusion of Cycles:} Assume, for the sake of contradiction, that the system does not terminate. This implies an infinite sequence of states $L_0, L_1, L_2, \dots$. Since the state space $\mathcal{L}$ is finite, this sequence must eventually revisit a state. That is, there must exist rounds $t_1 < t_2$ such that $L_{t_1} = L_{t_2}$.
    
    If the states are identical, their corresponding variance values must be equal: $V(L_{t_1}) = V(L_{t_2})$.
    
    However, the transition from $t_1$ to $t_2$ must involve at least one correction round (otherwise the state would have been stable and the sequence terminated). According to the monotonic descent property, any sequence of one or more corrections between $t_1$ and $t_2$ requires the variance to strictly decrease: $V(L_{t_1}) > V(L_{t_1+1}) > \dots > V(L_{t_2})$.
    
    This leads to the contradiction $V(L_{t_1}) = V(L_{t_2})$ and $V(L_{t_1}) > V(L_{t_2})$. Therefore, the initial assumption must be false. The system can never revisit a state.
\end{enumerate}
A trajectory through a finite state space that never revisits a state must be finite in length. The process must terminate. Termination occurs when the system reaches a fixed-point state $L^*$ where no further corrections are made (i.e., $\forall j, \alpha_{t,j} = 0$), and the system has converged.
\end{proof}
\section{Appendix: Details of Testing Protocol and Inplementation} \label{Appendix_protocol}

\subsection{Experimental Setup}
All experiments were conducted under controlled evaluation settings to ensure the comparability and reproducibility of results. For each benchmark:
\textbf{Sampling Consistency.} We fix the random seed and decoding parameters (e.g., temperature, top-$p$) across repeated runs. Each model's inference is repeated three times independently to estimate performance variance, with the average accuracy and standard deviation reported.
\textbf{Input Standardization.} Every model receives the same input format—limited to fields \texttt{question}, \texttt{options}, \texttt{answer}, and \texttt{answer\_idx}. No additional external knowledge or retrieval augmentation is provided unless explicitly tested (e.g., RAG baselines).
\textbf{Single-choice QA Format.} All evaluation items follow a single-best-choice multiple-choice format. A prediction is considered correct only if it exactly matches the gold answer index, with no need for free-text post-processing or alignment heuristics.
\textbf{Model Isolation.} No fine-tuning is performed on any of the large language models (LLMs). All LLMs use their original checkpoint and are served using \texttt{vLLM} (v0.7.2) for consistent inference latency and batching behavior.
\textbf{Hardware Environment.} All experiments are run on a dedicated 8$\times$A100-80GB GPU cluster. Each experiment is restricted to a single GPU node to ensure fair runtime comparison.
\textbf{Benchmark Integrity.} The benchmark order is shuffled between runs, and the LLM's internal sampling state is reset for each trial to avoid caching effects. We strictly follow the benchmark guidelines and do not alter test labels or answer keys.



\section{Appendix: Details of Experimental Environment} \label{Appendix_Environment}
The experiments were conducted on a high-performance computing system with robust hardware and software configurations to ensure efficient handling of large datasets and complex computations. The hardware setup included NVIDIA A100 GPUs with 80GB memory for accelerated computation, Intel Xeon Platinum 8358 CPUs with 64 cores operating at 1.20GHz for efficient multi-threaded processing, 1024GB of RAM, and 10TB of NVMe SSD storage for fast input/output operations.

The software environment was carefully configured for compatibility and reproducibility. The operating system used was Ubuntu 20.04 LTS (64-bit), and the primary deep learning framework was PyTorch (version 2.5.1), supplemented by TorchVision and Torchaudio extensions. The experiments were conducted using Python 3.10.16, with Conda (version 22.11.1) employed as the package manager to handle dependencies and virtual environments.

Key Python packages essential for the experiments included NumPy (2.0.1), and pandas (2.2.3) for data preprocessing and analysis. Visualization tasks were performed using Matplotlib (3.10.0). For deep learning tasks, PyTorch (2.5.1) served as the primary framework, and vLLM(v0.7.2) served as the LLM inference framework. 

A comprehensive list of installed Python packages is available upon request, capturing all dependencies required for reproducing the experiments. This configuration ensures the reported results are reproducible and highlights the environment's compatibility with the experimental setup.

\input{app_tab_medddx.tex}

\input{app_tab_medicalqa.tex}

%% file: app_tab_medddx.tex
\begin{table*}[t]
    \centering
    
        \vspace{-1.0em}
        \footnotesize
        \begin{tabular}{c|l|c||ccc|c}
            \toprule
                                                                                                                  & \multirow{3}{*}{\textbf{Method}} & \multirow{3}{*}{\textbf{Reference}} & \multicolumn{4}{c}{\textbf{MedDDx Benchmarks \citep{su2025kgarevion}}}                                                                                             \\
                                                                                                                  &                                  &                                     & \textbf{Basic}                                                         & \textbf{Intermediate}      & \textbf{Expert}             & \multirow{2}{*}{\textbf{AVE}}  \\
                                                                                                                  &                                  &                                     & Acc.($\pm$std)                                                         & Acc.($\pm$std)             & Acc.($\pm$std)              &                                \\ \midrule
            \multirow{3}{*}{{\begin{tabular}[c]{@{}c@{}}Graph \\Based\\ Methods\end{tabular}}}                    & QAGNN                            & NAACL2021                           & \underline{29.5($\pm$0.3)}                                             & \underline{26.5($\pm$0.2)} & \underline{25.3($\pm$0.3)}  & \underline{27.1}               \\
                                                                                                                  & JointLK                          & NAACL2022                           & 24.7($\pm$0.4)                                                         & 25.3($\pm$0.4)             & 24.4($\pm$0.4)              & 24.8                           \\
                                                                                                                  & Dragon                           & NeurIPS2022                         & 28.6($\pm$0.3)                                                         & 24.7($\pm$0.2)             & 24.0($\pm$0.4)              & 25.8                           \\ \midrule
            \multirow{4}{*}{{\begin{tabular}[c]{@{}c@{}}Multi \\ Agents \\ Methods\end{tabular}}}                 & MV-LLaMA3.1(8B)                  & -                                   & 39.6($\pm$1.0)                                                         & 32.8($\pm$0.6)             & 30.2($\pm$0.8)              & 34.2                           \\
                                                                                                                  & DyLAN                            & COLM2024                            & 39.3($\pm$1.5)                                                         & 33.5($\pm$0.9)             & 31.1($\pm$0.7)              & 34.6                           \\
                                                                                                                  & MedAgents                        & ACL2024                             & 41.0($\pm$0.7)                                                         & 35.7($\pm$1.1)             & 32.9($\pm$1.5)              & 36.5                           \\
                                                                                                                  & MDAgents                         & NeurIPS2024                         & \underline{42.1($\pm$1.3)}                                             & \underline{37.5($\pm$0.9)} & \underline{33.4($\pm$0.6)}  & \underline{37.7}               \\ \midrule
            \multirow{8}{*}{{\begin{tabular}[c]{@{}c@{}}General \\\&\\ Medical\\ LLMs\end{tabular}}}              & LLaMA2(7B)                       & Meta2023                            & 21.5($\pm$3.0)                                                         & 19.8($\pm$0.4)             & 19.2($\pm$1.2)              & 20.2                           \\
                                                                                                                  & Mistral(7B)                      & Mistral2023                         & 41.2($\pm$0.3)                                                         & 35.6($\pm$0.3)             & \underline{37.5($\pm$0.7)}  & \underline{38.1}               \\
                                                                                                                  & MedAlpaca(7B)                    & BHT2023                             & 39.9($\pm$1.2)                                                         & 32.5($\pm$0.4)             & 31.1($\pm$0.9)              & 34.5                           \\
                                                                                                                  & LLaMA2(13B)                      & Meta2023                            & 28.6($\pm$0.3)                                                         & 33.8($\pm$0.6)             & 31.7($\pm$0.6)              & 31.4                           \\
                                                                                                                  & PMC-LLaMA(7B)                    & SJTU2024                            & 8.7($\pm$1.5)                                                          & 8.6($\pm$0.2)              & 7.9($\pm$0.6)               & 8.4                            \\
                                                                                                                  & LLaMA3(8B)                       & Meta2024                            & 42.8($\pm$0.5)                                                         & 31.9($\pm$0.2)             & 30.6($\pm$0.9)              & 35.1                           \\
                                                                                                                  & Llama3-OB(8B)                    & Saama2024                           & 23.8($\pm$0.4)                                                         & 23.5($\pm$0.1)             & 22.9($\pm$2.0)              & 23.4                           \\
                                                                                                                  & \textbf{LLaMA3.1(8B)[baseline]}  & Meta2024                            & \underline{43.4($\pm$1.8)}                                             & \underline{36.8($\pm$0.2)} & 30.6($\pm$2.1)              & 36.9                           \\ \midrule
            \multirow{8}{*}{{\begin{tabular}[c]{@{}c@{}}General \\\&\\ Medical\\ LLMs \\with\\ COT\end{tabular}}} & CoT-LLaMA2(7B)                   & Meta2023                            & 28.9($\pm$1.0)                                                         & 26.5($\pm$0.6)             & 22.9($\pm$2.3)              & 26.1                           \\
                                                                                                                  & CoT-Mistral(7B)                  & Mistral2023                         & 40.4($\pm$1.0)                                                         & 36.8($\pm$2.3)             & \underline{37.9($\pm$2.7)}  & 38.4                           \\
                                                                                                                  & CoT-MedAlpaca(7B)                & BHT2023                             & 39.5($\pm$0.7)                                                         & 32.1($\pm$1.1)             & 31.2($\pm$1.0)              & 34.3                           \\
                                                                                                                  & CoT-LLaMA2(13B)                  & Meta2023                            & 25.6($\pm$1.3)                                                         & 26.3($\pm$1.0)             & 24.3($\pm$1.6)              & 25.4                           \\
                                                                                                                  & CoT-PMC-LLaMA(7B)                & SJTU2024                            & 8.8($\pm$0.2)                                                          & 7.7($\pm$0.4)              & 6.3($\pm$0.5)               & 7.6                            \\
                                                                                                                  & CoT-LLaMA3(8B)                   & Meta2024                            & \underline{43.4($\pm$0.9)}                                             & 36.8($\pm$0.4)             & 31.3($\pm$0.3)              & 37.2                           \\
                                                                                                                  & CoT-Llama3-OB(8B)                & Saama2024                           & 37.0($\pm$1.3)                                                         & 33.0($\pm$0.1)             & 32.7($\pm$1.1)              & 34.2                           \\
                                                                                                                  & CoT-LLaMA3.1(8B)                 & Meta2024                            & {43.9($\pm$1.7)}                                                       & \underline{39.3($\pm$0.5)} & 32.2($\pm$1.4)              & \underline{ 38.5 }             \\ \midrule
            \multirow{4}{*}{{\begin{tabular}[c]{@{}c@{}}RAG \\\&\\ Based\\ Methods\end{tabular}}}                 & Self-RAG(7B)                     & ICLR2024                            & 23.8($\pm$0.7)                                                         & 19.9($\pm$3.7)             & 22.4($\pm$4.5)              & 22.0                           \\
                                                                                                                  & Self-RAG(13B)                    & ICLR2024                            & 24.9($\pm$1.0)                                                         & 29.0($\pm$1.8)             & 26.6($\pm$3.1)              & 26.8                           \\
                                                                                                                  & KG-Rank(13B)                     & ACL-w2024                           & 25.3($\pm$2.1)                                                         & 25.6($\pm$1.3)             & 23.4($\pm$1.0)              & 24.8                           \\
                                                                                                                  & MedRAG(70B)                      & Oxon2024                            & \underline{36.5($\pm$0.8)}                                             & \underline{34.8($\pm$1.1)} & \underline{ 32.7($\pm$0.3)} & \underline{ 34.7 }             \\ \midrule
            \multirow{1}{*}{{\begin{tabular}[c]{@{}c@{}}Logic Based\end{tabular}}}                                & MedLA+LLaMA3.1(8B)               & Ours                                & \textbf{48.2($\pm$1.2)}                                                & \textbf{43.0($\pm$2.1)}    & \textbf{41.7($\pm$0.8)}     & \textbf{44.3 ($\uparrow$ 7.4)} \\
            \bottomrule
        \end{tabular}
      \caption{\textbf{The performance of our MedLA on MedDDx Benchmarks.} The table includes the accuracy along with the standard deviation($\pm$std) for each metric. The results demonstrate the effectiveness of MedLA in addressing complex medical reasoning tasks across different datasets. OB means OpenBioLLM, MV means Majority Voting, and AVE means averaged accuracy. The best results are highlighted in \textbf{bold}, while the best results of each method's block are marked with an \underline{underline}. Reference indicates the paper where the method was first introduced. The results of the baselines are taken from \citep{su2025kgarevion}.}\label{tab_main_results_medddx}
\end{table*}

%% file: app_tab_medicalqa.tex
\begin{table*}[t]
    \centering
  
    \footnotesize
    \vspace{-0.1cm}
        \begin{tabular}{c|l|c||ccc|c} \toprule
                                                                                                                 & \multirow{3}{*}{\textbf{Method}} & \multirow{3}{*}{\textbf{Reference}} & \multicolumn{4}{c}{\textbf{Multi-choice medical QA benchmarks \citep{xiong2024medrag}}}                                                                                              \\
                                                                                                                 &                                  &                            & MMLU-Med                                                                              & MedQA-US                    & BioASQ-Y/N                  & \multirow{2}{*}{\textbf{AVE}}          \\
                                                                                                                 &                                  &                            & Acc.($\pm$std)                                                                        & Acc.($\pm$std)              & Acc.($\pm$std)              &                                \\ \midrule
            \multirow{3}{*}{{\begin{tabular}[c]{@{}c@{}}Graph \\ Based \\ Methods\end{tabular}}}                 & QAGNN                            & NAACL2021                  & 31.7($\pm$0.6)                                                                        & 47.0($\pm$0.3)              & \underline{70.7($\pm$0.6)}  & 49.8                           \\
                                                                                                                 & JointLK                          & NAACL2022                  & 28.8($\pm$0.6)                                                                        & 42.5($\pm$0.2)              & 70.6($\pm$0.5)              & 47.3                           \\
                                                                                                                 & Dragon                           & NeurIPS2022                & \underline{31.9($\pm$0.3)}                                                            & \underline{47.5($\pm$0.2)}  & 70.6($\pm$0.3)              & \underline{50.0}               \\ \midrule
            \multirow{4}{*}{{\begin{tabular}[c]{@{}c@{}}Multi \\ Agents \\ Methods\end{tabular}}}                & MV-LLaMA3.1(8B)                   & -                          & 60.2($\pm$0.5)                                                                        & 46.8($\pm$0.4)              & \underline{65.2($\pm$0.4)}  & 57.4                           \\
                                                                                                                 & DyLAN                            & COLM2024                   & 62.5($\pm$0.3)                                                                        & 51.6($\pm$0.6)              & 63.8($\pm$0.5)              & 59.3                           \\
                                                                                                                 & MedAgents                        & ACL2024                    & 64.3($\pm$0.4)                                                                        & 53.2($\pm$0.3)              & 64.1($\pm$0.6)              & 60.5                           \\
                                                                                                                 & MDAgents                         & NeurIPS2024                & \underline{65.0($\pm$0.2)}                                                            & \underline{53.4($\pm$0.2)}  & 64.0($\pm$0.3)              & \underline{60.8}               \\ \midrule
            \multirow{8}{*}{{\begin{tabular}[c]{@{}c@{}}General \\ \&\\ Medical\\ LLMs\end{tabular}}}            & LLaMA2(7B)                        & Meta2023                   & 37.6($\pm$0.6)                                                                        & 28.1($\pm$0.4)              & 56.8($\pm$0.6)              & 40.8                           \\
                                                                                                                 & Mistral(7B)                       & Mistral2023                & 63.4($\pm$0.4)                                                                        & 47.7($\pm$0.7)              & 64.4($\pm$0.1)              & 58.5                           \\
                                                                                                                 & MedAlpaca(7B)                     & BHT2023                    & 60.0($\pm$0.4)                                                                        & 40.1($\pm$0.1)              & 49.3($\pm$3.4)              & 49.8                           \\
                                                                                                                 & LLaMA2(13B)                       & Meta2023                   & 44.2($\pm$0.6)                                                                        & 25.3($\pm$0.4)              & 34.6($\pm$0.5)              & 34.7                           \\
                                                                                                                 & PMC-LLaMA(7B)                     & SJTU2024                   & 20.7($\pm$1.1)                                                                        & 24.7($\pm$0.4)              & 34.6($\pm$1.7)              & 26.7                           \\
                                                                                                                 & LLaMA3(8B)                        & Meta2024                   & 63.4($\pm$0.5)                                                                        & {56.6($\pm$0.4)}            & 65.4($\pm$0.6)              & 61.8                           \\
                                                                                                                 & LLaMA3-OB(8B)                     & Saama2024                  & 63.6($\pm$0.6)                                                                        & 38.3($\pm$0.3)              & 62.2($\pm$0.3)              & 54.7                           \\
                                                                                                                 & \textbf{LLaMA3.1(8B)[baseline]}          & Meta2024                   & \underline{67.7($\pm$0.7)}                                                            & \underline{56.3($\pm$0.6)}  & \underline{68.7($\pm$0.6)}  & \underline{64.2}               \\ \midrule
            \multirow{8}{*}{{\begin{tabular}[c]{@{}c@{}}General \\ \&\\ Medical\\ LLMs \\with COT\end{tabular}}} & COT-LLaMA2(7B)                    & Meta2023                   & 31.8($\pm$0.5)                                                                        & 25.1($\pm$0.2)              & 54.7($\pm$1.1)              & 37.2                           \\
                                                                                                                 & COT-Mistral(7B)                   & Mistral2023                & 63.4($\pm$0.3)                                                                        & 47.4($\pm$0.2)              & 65.1($\pm$0.2)              & 58.6                           \\
                                                                                                                 & COT-MedAlpaca(7B)                 & BHT2023                    & 60.3($\pm$0.4)                                                                        & 39.9($\pm$0.3)              & 48.5($\pm$2.5)              & 49.6                           \\
                                                                                                                 & COT-LLaMA2(13B)                   & Meta2023                   & 41.5($\pm$0.5)                                                                        & 35.4($\pm$0.6)              & 36.3($\pm$1.3)              & 37.7                           \\
                                                                                                                 & COT-PMC-LLaMA(7B)                 & SJTU2024                   & 20.4($\pm$0.8)                                                                        & 20.8($\pm$0.2)              & 20.8($\pm$0.6)              & 20.7                           \\
                                                                                                                 & COT-LLaMA3(8B)                    & Meta2024                   & 65.1($\pm$0.5)                                                                        & \underline{55.2($\pm$0.3)}  & 64.2($\pm$0.5)              & 61.5                           \\
                                                                                                                 & COT-Llama3-OB(8B)                 & Saama2024                  & 57.1($\pm$0.3)                                                                        & 39.5($\pm$0.2)              & 64.6($\pm$0.6)              & 53.7                           \\
                                                                                                                 & COT-LLaMA3.1(8B)                  & Meta2024                   & \underline{68.1($\pm$0.5)}                                                            & 54.9($\pm$0.3)              & \underline{70.6($\pm$0.5)}  & \underline{64.5 }              \\\midrule
            \multirow{4}{*}{{\begin{tabular}[c]{@{}c@{}}RAG \\ Based\\ Methods\end{tabular}}}                    & Self-RAG (7B)                    & ICLR2024                   & 32.2 ($\pm$1.9)                                                                       & 38.0 ($\pm$2.8)             & 59.4 ($\pm$1.2)             & 43.2                           \\
                                                                                                                 & Self-RAG (13B)                   & ICLR2024                   & 50.2 ($\pm$0.4)                                                                       & 40.8 ($\pm$2.0)             & 64.6 ($\pm$5.0)             & 51.9                           \\
                                                                                                                 & KG-Rank (13B)                    & ACL-w2024                  & 45.2 ($\pm$0.5)                                                                       & 36.2 ($\pm$1.1)             & 50.3 ($\pm$1.5)             & 43.9                           \\
                                                                                                                 & MedRAG (70B)                     & Oxon2024                   & \underline{57.9 ($\pm$1.5)}                                                           & \underline{48.7 ($\pm$1.4)} & \underline{71.9 ($\pm$1.8)} & \underline{59.5}               \\
            \midrule
            \multirow{1}{*}{{\begin{tabular}[c]{@{}c@{}}Logic Based \end{tabular}}}                      & MedLA + LLaMA3.1(8B)              & Ours                       & \textbf{70.7($\pm$0.1)}                                                               & \textbf{62.6($\pm$0.1)}     & \textbf{76.5($\pm$0.1)}     & \textbf{69.9($\uparrow$5.7)} \\
            \bottomrule
        \end{tabular}
     \caption{\textbf{The performance of our proposed MedLA model on Multi-choice medical QA benchmarks \citep{xiong2024medrag}.} The table includes the accuracy (Acc) along with the standard deviation ($\pm$std) for each metric. The results demonstrate the effectiveness of MedLA in addressing complex medical reasoning tasks across different datasets. } \label{tab_results_medicalqa}
    \vspace{-1.5em}
\end{table*}